\documentclass[]{aastex631}
\usepackage{amsmath,amssymb}
\usepackage{graphicx}
\usepackage{graphics}
\usepackage{dsfont}
\usepackage{dcolumn}
\usepackage{bm}
\usepackage{blindtext}
\usepackage{stackengine}
\usepackage{mathptmx}
\usepackage{etoolbox}
\usepackage{color}
\usepackage{epstopdf} 
\usepackage{caption}
\usepackage{subcaption}
\definecolor{bluepig}{rgb}{0.2, 0.2, 0.6}
\definecolor{bluencs}{rgb}{0.0, 0.53, 0.74}
\definecolor{darkcerulean}{rgb}{0.03, 0.27, 0.49}
\definecolor{brightpink}{rgb}{1.0, 0.0, 0.5}
\newcommand{\markup}[1]{\textcolor{black}{#1}}

\newcommand{\Losseqn}{\mathcal{L}_{eqn}}
\newcommand{\LossBC}{\mathcal{L}_{BC}}
\newcommand{\Lossini}{\mathcal{L}_{ini}}
\newcommand{\Lossdata}{\mathcal{L}_{data}}

\usepackage{pifont}
\newcommand{\cmark}{\ding{51}}%
\newcommand{\xmark}{\ding{55}}%
\usepackage{lipsum}
\usepackage{natbib}

\setcitestyle{authoryear,open={(},close={)}}

\graphicspath{{./}}
\begin{document}

\title{Physics-informed neural networks for modeling dynamic linear elasticity}

\author{Vijay Kag}
\affiliation{Bosch Research and Technology Center - Bangalore, India}

\author{Venkatesh Gopinath}
 \altaffiliation{Corresponding author \\
 \textit{email address}: gopinath.venkatesh2@in.bosch.com \\ \\ \\ \\
 \noindent\textit{{Preprint}}}
\affiliation{Bosch Research and Technology Center - Bangalore, India
}%

\begin{abstract}

In this work, we present the physics-informed neural network (PINN) model applied particularly to dynamic problems in solid mechanics. We focus on forward and inverse problems. Particularly, we show how a PINN model can be used efficiently for material identification in a dynamic setting. In this work, we assume linear continuum elasticity. We show results for two-dimensional (2D) plane strain problem and then we proceed to apply the same techniques for a three-dimensional (3D) problem. As for the training data we use the solution based on the finite element method. We rigorously show that PINN models are accurate, robust and computationally efficient, especially as a surrogate model for material identification problems. Also, we employ state-of-the-art techniques from the PINN literature which are an improvement to the vanilla implementation of PINN. Based on our results, we believe that the framework we have developed can be readily adapted to computational platforms for solving multiple dynamic problems in solid mechanics.

\end{abstract}


\section{Introduction}
Dynamic problems in solid mechanics are ever-present in industrial settings across multiple domains such as automotive, ship building, aerospace to name a few. Traditionally, experiments or measurements are conducted to analyze these problems. In the event that experiments are practically unfeasible to perform or in the case of repetitive need for it which is more often the case, numerical methods had come to the forefront for obtaining solutions for these problems for many decades. For example the finite element method (FEM) \citep{zeinkiewicz_book} has been one of the most useful numerical techniques to solve many problems in the area of solid mechanics. However, in many situations, for example, in material parameter identification problems, numerical simulations incur high computational costs. Nevertheless, they are extensively used until the simulation time is beyond an acceptable threshold. So, the practical question of whether such problems can be solved otherwise is a motivation to seek newer ideas and methods. One such area of recent interest is the rapidly growing field of deep learning \citep{Lacun_nature2015} which is based on the availability of large data. Deep learning has had great impact on fields such as computer vision \citep{elsayed2018,Liu2022}, autonomous driving \citep{grigorescu2019,fujiyoshi2019}, healthcare \citep{miotto2018,esteva2018} among many. There has also been quite some development on deep learning based research for solving problems in mechanics \citep{brunton2020,kutz2017,sun2021}. 

Of particular interest in this work is their use in problems based on solid mechanics. Although deep learning is seeing a recent upsurge in applications due to the availability of advanced computing resources, it has been used as early as the late $1980$s for problems within the context of building structural performance assessment  \citep{adeli1989}. Deep learning has also been used multiple times for constitutive modeling in solids over the years \citep{ghaboussi,mozaffar}. Constitutive modeling aims to construct a mathematical form of the relationship between the applied stress to the material and the resulting strain. Many a time, for complex materials or assembly of multiple components, there are gaps in the constitutive models and deep learning based on data addresses these shortcomings. Recent advances in deep learning are also used towards structural health monitoring \citep{sun2021,yoon2022}. 

Usually, deep learning approaches use a lot of data and they don't account for underlying physical laws of the system under study. So, they may be unfeasible to use if there is a dearth of data. One such method which circumvents this issue while using deep learning is the PINN. It uses the information of governing equations of the system along with the boundary and initial conditions for constructing the overall loss function which the neural network tries to minimise. It has an effect of penalization of nonphysical solutions. It was introduced in the seminal work of \citep{raissi19}. Since then, there has been a plethora of research works on its development \citep{ameya2020,hfm20,nsfnets21,wang20,augpinns} and applications \citep{lucor22,cai2021,ameya2022,vijay2022}. We are interested in this work in the use of PINN models for solving problems in solid mechanics. One of the pioneering works on this topic was done by \citep{raissi2021}. They provide a framework for the use of PINN in the context of forward, inverse and surrogate modeling. Since then, there has been significant applications of PINN in this area. For example, \citep{karniadakis2020} uses PINN to identify nonhomogeneous hyperelastic materials. \citep{wessels2023} perform material identification from both experimental data as well as full-field displacement data with noise. In the work of \citep{wu2023}, they apply PINN for identifying material properties in linear elasticity. Another important work was done by \citep{haghighat2023} where they use PINN to characterize constitutive models for elastoplastic solids. \citep{goswami2022} applied PINN to discover constitutive models in problems involving elastic and viscoplastic solids. Recently, \citep{roy2023} also provide a PINN framework for solving linearly elastic problems such as an end-loaded cantilever beam and Kirchoff-Love thin plate. Also, \citep{royguha2023} use PINN where they combine data and physics to solve problems in plasticity. PINN has also been used to detect defects in materials in the work of \citep{zhang2022}. Recently, use of PINN for complex geometries is shown by \citep{perdikaris2022} where they apply it on a hyperelastic problem. Aside from PINN, in the context of solid mechanics, in \citep{cueto2019}, they use similar ideas of combining data and physical laws of thermodynamics to correct existing constitutive models for problems in hyperelasticity.  

While there has been reasonable amount of work on PINN applied to solid mechanics, one of the questions which still persists is whether PINN can be successfully used for a realistic problem in solid mechanics which is dynamic in nature where the acceleration terms in the governing equations are active.  In this work, we use PINN to gauge it's applicability and effectiveness for dynamic forward and inverse problems in a linearly elastic setting. For training data, we use the FEM simulation results. Of particular focus in this study is the use of PINN to create a surrogate model for material identification. This, in particular, is a major motivation coming from an industrial perspective because there are multiple areas in the industry where solutions to such problems are computationally expensive and we show that the technology of PINN can circumvent this issue. 

The study is organized as follows: in Sec. \ref{Sec:PINN}, the methodology of PINN is introduced; Following that, the governing equations for a linearly elastic problem is discussed in Sec. \ref{Sec:LinEla}; In Sec. \ref{Sec:PINN_LinEla}, an adaptation of a PINN model to the linearly elastic problem is shown along with a discussion on the implementation; Next, in Sec. \ref{Sec:Results}, the cantilever problem is introduced and subsequently, the forward, inverse and the surrogate modeling results are explored in detail with an emphasis on industrial applicability. Also, in the subsection Sec. \ref{Sec:3D}, the extension of the results to a three-dimensional setting is explored. Finally, the conclusions are discussed in Sec. \ref{Sec:Conclusion}. 



\section{Physics-informed neural networks}
\label{Sec:PINN}

In this section, we will briefly describe the technology of PINN based on the seminal article \citep{raissi19}. A general form of a governing equation describing a physical process can be given as,
\begin{align}
    \partial_t {\bf u} = \mathcal{M} {\bf u} + \mathcal{N} ( {\bf u} ),
    \label{eqn:Gen_NS}
\end{align}
where $\mathcal{N}$ is the nonlinear operator and $\mathcal{M}$ is the linear operator. Using the neural network graph structure \citep{optbook}, the linear and nonlinear terms are calculated. For accurate neural network based approximation of the solution, a loss function has to be minimized \citep{raissi19,nsfnets21}. The loss functions are defined as follows.

The equation loss $\Losseqn$ is given as: 
\begin{equation}
    \Losseqn = \frac{1}{N_{eqn}} \sum_{n=1}^{N_{eqn}} |\partial_t {\bf u} ({\bf x}_n^e, t_n^e) - \mathcal{M} {\bf u} ({\bf x}_n^e, t_n^e) - \mathcal{N}({\bf u} ({\bf x}_n^e, t_n^e))|^2,
    \label{eqn:loss_eqn}
\end{equation}
where $({\bf x}^e, t^e)$ denote the collocation points where equation loss is calculated and $N_{eqn}$ denotes the total number of these collocation points. The boundary and initial condition loss functions are given as: 

\begin{align}
    \LossBC  = & \frac{1}{N_{BC}} \sum_{n = 1}^{N_{BC}} |G ( {\bf u} ({\bf x}^b_n, t^b_n) )|^2, \\
    \Lossini = & \frac{1}{N_{ini}} \sum_{n = 1}^{N_{ini}} |{\bf u} ( {\bf x}^i_n, 0 ) - {\bf u}^e ({\bf x}^i_n, 0)|^2,  \label{eqn:loss_ini}
\end{align}
where $({\bf x}^b, t^b)$ are collocation points on the boundaries where the boundary loss is trained and $({\bf x}^i, 0)$ are collocation points where the initial condition loss is trained. 
$N_{BC}$ denotes the number of collocation points on the boundaries and $N_{ini}$ denotes the number of collocation points at initial time. ${\bf u}^e ({\bf x}, 0)$ is the initial conditions for the set of equations in \eqref{eqn:Gen_NS}. 
{\markup{The function $G$ is a measure of the deviation of the predicted solution from the imposed boundary conditions (BCs). It takes only positive real values for any solution field.}}
We also compute the loss data $\Lossdata$ which measures the deviation of the predicted solution from the exact solution. 
It is defined as,
\begin{align}
\Lossdata = \frac{1}{N_{data}} \sum_{n=1}^{N_{data}}  |{\bf u} ({\bf x}^d_n, t^d_n) - {\bf u}^{e} ({\bf x}^d_n, t^d_n)|^2, \label{eqn:loss_data}
\end{align}
where $({\bf x}^d, t^d)$ are collocation points for training the system on a sparse set of interior points and ${\bf u}^{e}$ denotes the exact solution which is obtained either from simulation or from experimental data. 
$N_{data}$ is the number of collocation points for training on data of interior points. In this work, we consider only the data loss and the equation loss. The total loss is thus given by:
\begin{align}
    \mathcal{L}_{tot} = \lambda_1 \Lossdata + \lambda_2 \Losseqn,
    \label{eqn:total_loss}
\end{align}

{\markup{where}} $\lambda_1$ and $\lambda_2$ are weight coefficients for the contribution from data and equation losses respectively. 

The inputs to the neural network are the spatial and temporal coordinates $({\bf x},t)$. According to \cite{goodfellow}, each network layer output is given as: 
\begin{align}
    {\bf z}_l = \sigma_l \left( {\bf z}_{l-1} {\bf W}_l + {\bf b}_l \right), 
\end{align}

where $l$ is the layer number, ${\bf W}$ represents the weight, ${\bf b}$ the bias and $\sigma_l$ is the activation function. The weights and biases are optimized so as to minimize the total loss $L_{tot}$. 

\section{Linear elasticity}
\label{Sec:LinEla}
In this section, we give an overview of the governing equations for linear elasticity. We start with the equation for dynamic momentum balance 

\begin{align}
    \nabla \cdot \bm{\sigma(\bm{x})} + \rho \bm{b(\bm{x})} = \rho \frac{\partial^2 \bm{u(\bm{x})}}{\partial t^2}, \ \bm{x} \in \Omega,
    \label{maineqn}
\end{align}
where, $\bm{\sigma}$ is the stress tensor field, $\rho$ is the density of the material, $\bm{b}$ is the body force and $\bm{u}$ is the displacement. $\bm{x}$ are the points inside the domain $\Omega$. The stress tensor field is related to the strain tensor field $\bm{\epsilon}(\bm{x})$ by a constitutive relation

\begin{align}
\bm{\sigma} = \lambda \text{tr}(\bm{\epsilon}(\bm{x})) \mathds{1} + 2 \mu \bm{\epsilon}(\bm{x}), \ \bm{x} \in \Omega,
\label{consteqn}
\end{align}
where $\lambda$ and $\mu$ are the Lam\'e constants and $\mathds{1}$ is the second order identity tensor. Also, the strain tensor field is related to the displacement vector field $\bm{u}(\bm{x})$ by the kinematic relation

\begin{align}
\bm{\epsilon}(\bm{x}) = \frac{1}{2} (\nabla \bm{u}(\bm{x}) + \nabla \bm{u}^T(\bm{x})), \ \bm{x} \in \Omega.
\label{kinemeqn}
\end{align}
We assume plane-strain conditions and no body forces. Therefore, in scalar form the Eq. \eqref{maineqn} becomes 

\begin{align}
\frac{\partial \sigma_{xx}}{\partial x} + \frac{\partial \sigma_{xy}}{\partial y} = \rho \frac{\partial^2 u_x}{\partial t^2}, \label{maineqn1}\\
\frac{\partial \sigma_{yy}}{\partial y} + \frac{\partial \sigma_{xy}}{\partial x} = \rho \frac{\partial^2 u_y}{\partial t^2}.
\label{maineqn2}
\end{align}
Likewise, the constitutive relation in Eq. \eqref{consteqn} based on the kinematic relation from Eq. \eqref{kinemeqn} becomes

\begin{align}
    \sigma_{xx} = (\lambda + 2 \mu) \frac{\partial u_x}{\partial x} + \lambda \frac{\partial u_y}{\partial y}, \label{consteqn1}\\
    \sigma_{yy} = (\lambda + 2 \mu) \frac{\partial u_y}{\partial y} + \lambda \frac{\partial u_x}{\partial x}, \label{consteqn2}\\
     \sigma_{xy} = \mu \biggl( \frac{\partial u_x}{\partial y} + \frac{\partial u_y}{\partial x} \biggr). \label{consteqn3}
\end{align}
Next, we scale the quantities in these equations as shown below

\begin{align}
    x^* = \frac{x}{x_c}, \ y^* = \frac{y}{y_c}, \ u_x^* = \frac{u_x}{{u_x}_c}, \ u_y^* = \frac{u_y}{{u_y}_c}, \ \sigma_{xx}^* = \frac{\sigma_{xx}}{{\sigma_{xx}}_c}, \ \sigma_{yy}^* = \frac{\sigma_{yy}}{{\sigma_{yy}}_c}, \ \sigma_{xy}^* = \frac{\sigma_{xy}}{{\sigma_{xy}}_c}, \ t^* = \frac{t}{t_c}, \ \lambda^* = 
    \frac{\lambda}{\lambda_c}, \ \mu^* = \frac{\mu}{\mu_c}, \ \rho^* = \frac{\rho}{\rho_c},
    \label{scalings}
\end{align}
where the scaling factors $\boldsymbol{\cdot}_c$ are chosen to be the absolute maximum value of the quantities under consideration for the forward problem. We assume $x_c = y_c = l_c$ and $\lambda_c = \mu_c$. For forward simulations, we know the value of the material properties. However, for the inverse problem or surrogate modeling, since we do not have any prior information on the material properties, we have to assume a value for $\lambda_c$ and $\mu_c$. In a practical scenario, for an inverse problem, for training data, one will perform measurements on a newly manufactured material and one would know a rough value for the properties which can then be chosen for the scaling. Also, while performing neural network prediction, the quantities with $\boldsymbol{\cdot}^*$ will be scaled back to original range. 

\section{PINN model for linear elasticity}
\label{Sec:PINN_LinEla}

\begin{figure*}
\includegraphics[width=\textwidth,height=\textheight,keepaspectratio]{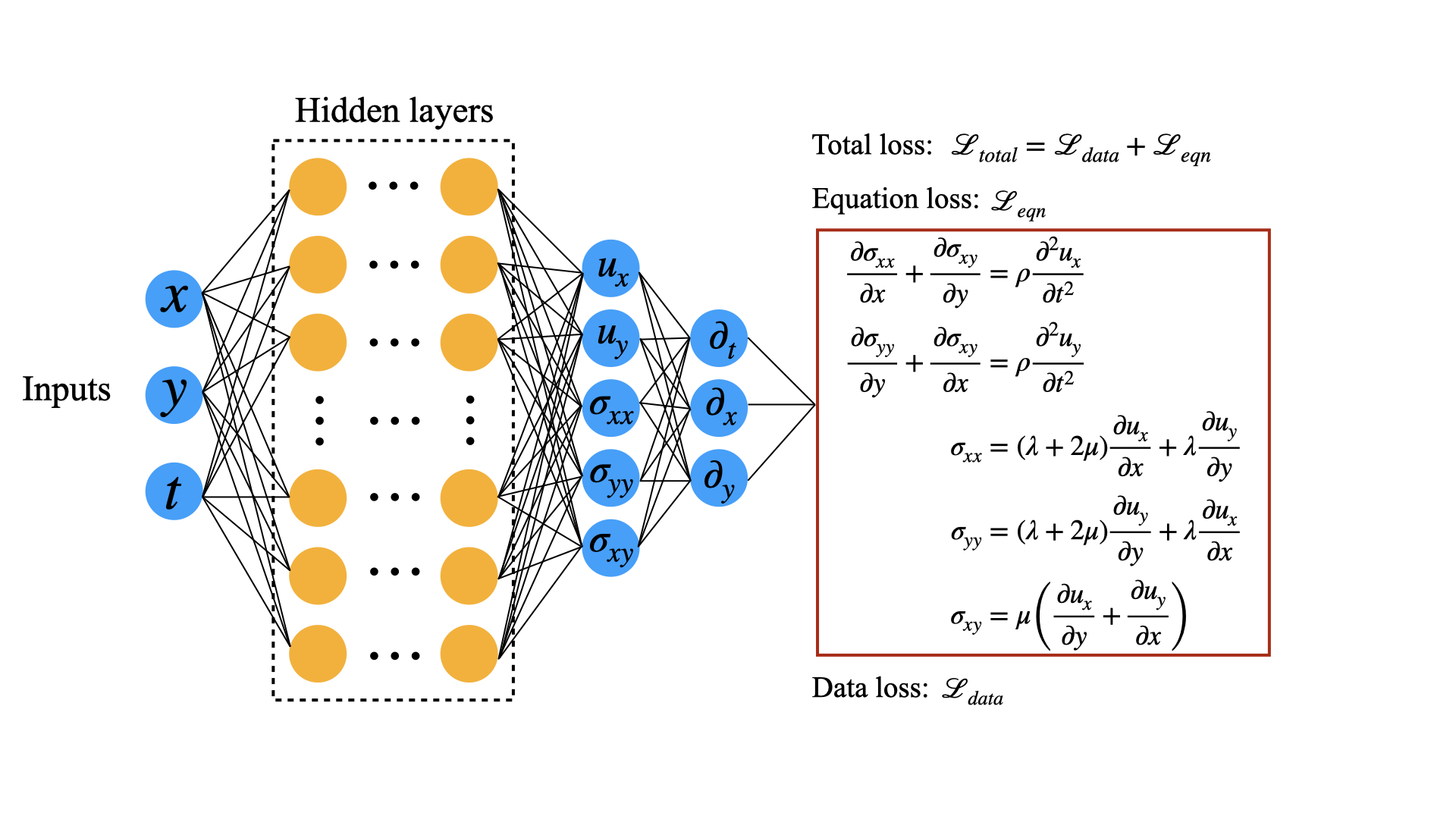}
\caption{\label{sch_pinn} Physics-informed neural network layout}
\end{figure*}

We apply the PINN framework to the problem based on linearly elastic solid mechanics. As for the neural network architecture, we use a single-network model where it takes the inputs $(x^*,y^*,t^*)$ and outputs the quantities of interest $(u_x^*,u_y^*,\sigma_{xx}^*,\sigma_{yy}^*,\sigma_{xy}^*)$. The equation loss is constructed based on the strong form of the governing equations provided in Eqs. (\ref{maineqn1},\ref{maineqn2}) and Eqs. (\ref{consteqn1},\ref{consteqn2},\ref{consteqn3}). The proposed neural network architecture is shown in Fig. \ref{sch_pinn}. We use a single-network architecture where input to the last layer is identical for all outputs. For clarity, we re-write Eq. \eqref{eqn:total_loss}

\begin{align}
\mathcal{L}_{tot} = \Lossdata +  \Losseqn,
\end{align}

where the individual losses based on Eqs. \eqref{eqn:loss_eqn} and \eqref{eqn:loss_data} can be expanded as
\begin{align}
\Lossdata =  \|{u_x^*} - {u_x^*}^{d}\|_{data} + \|{u_y^*} - {u_y^*}^{d}||_{data} + \|{\sigma_{xx}^*} - {\sigma_{xx}^*}^{d}\|_{data} + \|{\sigma_{yy}^*} - {\sigma_{yy}^*}^{d}\|_{data} + \|{\sigma_{xy}^*} - {\sigma_{xy}^*}^{d}\|_{data},\label{eqn:loss_data_expand} 
\end{align}

\begin{equation}
\begin{split}
{  \mathcal{L}_{eqn} = \alpha_1 \mathcal{L}_{x} + \alpha_2 \mathcal{L}_{y} + \alpha_3 \mathcal{L}_{xx} +
\alpha_4 \mathcal{L}_{yy} + \alpha_5 \mathcal{L}_{xy} 
}
\end{split}
\label{loss_eqn_alpha}
\end{equation}

\begin{equation}
\begin{split}
{
\mathcal{L}_{x} =  \biggr\| \frac{\partial \sigma_{xx}^*}{\partial x^*} +   \frac{{\sigma_{xy}}_c}{{\sigma_{xx}}_c}  \frac{\partial \sigma_{xy}^*}{\partial y^*} - \frac{ {u_x}_c \rho_c  l_c}{ {\sigma_{xx}}_c t_c^2 }  \rho^* \frac{\partial^2 u_x^*}{\partial {t^*}^2}  \biggr\|_{eqn}
 }
\end{split}
\end{equation}

\begin{equation}
\begin{split}
{
\mathcal{L}_{y} =  \biggr\|   \frac{{\sigma_{xy}}_c}{{\sigma_{xx}}_c} \frac{\partial \sigma_{xy}^*}{\partial x^*} + \frac{{\sigma_{yy}}_c}{{\sigma_{xx}}_c}  \frac{\partial \sigma_{yy}^*}{\partial y^*}  - \frac{ {u_y}_c \rho_c  l_c}{ {\sigma_{xx}}_c t_c^2 }  \rho^* \frac{\partial^2 u_y^*}{\partial {t^*}^2}  \biggr\|_{eqn}
 }
\end{split}
\end{equation}

\begin{equation}
\begin{split}
{
\mathcal{L}_{xx} =\biggr\|  \frac{ {\sigma_{xx}}_c l_c }{ {u_y}_c \lambda_c}  \sigma_{xx}^* - \frac{ {u_x}_c }{ {u_y}_c } (\lambda^* + 2 \mu^*) \frac{\partial u_x^* }{\partial x^*} - \lambda^* \frac{\partial {u_y}^*}{\partial y^*} \biggr\|_{eqn}
}  
\end{split}
\end{equation}

\begin{equation}
\begin{split}
{
\mathcal{L}_{yy} =\biggr\|  \frac{ {\sigma_{yy}}_c l_c }{ {u_y}_c \lambda_c}  \sigma_{yy}^*  -  \frac{ {u_x}_c }{ {u_y}_c } \lambda^* \frac{\partial u_x^*}{\partial x^*} -  (\lambda^* + 2 \mu^*) \frac{\partial u_y^* }{\partial y^*} \biggr\|_{eqn}
}  
\end{split}
\end{equation}

\begin{equation}
\begin{split}
{
\mathcal{L}_{xy} =\biggr\|  \frac{ {\sigma_{xy}}_c l_c }{ {u_y}_c \lambda_c}  \sigma_{xy}^* -  \mu^*\biggl( \frac{\partial u_y^* }{\partial x^*} + \frac{ {u_x}_c }{{u_y}_c} \frac{\partial u_x^*}{\partial y^*} \biggr)  \biggr\|_{eqn}  
}  
\end{split}
\end{equation}

The norm $\|.\|_{data} = \frac{1}{N_{data}} \sum_{n=1}^{N_{data}}  |{q} ({\bf x}^d_n, t^d_n)|^2$ where $q$ is a quantity of interest. Likewise, $\|.\|_{eqn} = \frac{1}{N_{eqn}} \sum_{n=1}^{N_{eqn}}  |{q} ({\bf x}^e_n, t^e_n)|^2$. Also, $\alpha_1$ to $\alpha_5$ in front of the terms in Eq. \ref{loss_eqn_alpha} are the weight coefficients. For the neural network training, we consider couple of useful suggestions provided by \citep{wang2023}. First, we use the modified neural network architecture proposed by \citep{wang2021} where the forward pass of a $L$-layer architecture is defined as
\begin{align}
    \bm{u}&= \sigma (\bm{x} \bm{w}_1 + \bm{b}_1), \bm{v} = \sigma (\bm{x} \bm{w}_2 + \bm{b}_2), \\
    \bm{h}_1 &= \sigma(\bm{x} \bm{w}_l + \bm{b}_l), \\
    \bm{z}_l &= \sigma(\bm{h}_l \bm{w}_l + \bm{b}_l), \ l=1,....,L-1, \\
    \bm{h}_{l+1} &= (1-\bm{z}_l) \odot \bm{u} + \bm{z}_l \odot \bm{v}, \ l=1,....,L-1,
\end{align}
where $\odot$ denotes element-wise multiplication. The network output is given as
\begin{align}
    \bm{u}(\bm{x}) = \bm{h}_l \bm{w}_l + \bm{b}_l,
\end{align}
and the trainable parameters are then
\begin{align}
    \bm{\theta} = {\bm{w}_1, \bm{b}_1, \bm{w}_2, \bm{b}_2, (\bm{w}_l, \bm{b}_l)^L_{l=1}}.
\end{align}

For optimization, we use stochastic gradient descent using the \textit{Adam} optimizer \citep{adam}. Other important parameters for our training are the batch-size, epochs, number of equation points and data points. The batch-size is the number of training instances in the batch and one epoch is one complete training round covering the entire data-set. For example, if we have a data-set of size $N$ and a batch-size of $N_{batch}$, then the total number of batches will be $N/N_{batch}$. Therefore, for one epoch, we do the gradient descent optimization for each batch. Also, in each gradient descent step, the equation points are randomly chosen from the domain $\Omega$. Hence training through one epoch consists of  $N/N_{batch}$ number of stochastic gradient descent steps. Unless specified, for all 2D PINN simulations, we keep the equation loss coefficients $(\alpha_1, \alpha_2,\alpha_3, \alpha_4, \alpha_5)$ as unity.

The other suggestion from \citep{wang2023} we consider is the non-dimensionalization of the governing equations based on the scalings provided in Eq. \eqref{scalings}. This ensures that the output quantities of interest from the neural network are within reasonable range of values. Our scalings are similar to the work of \citep{henkes2022}. Since we use data loss as one of the cost functions to minimize, we have access to the maximum values of each quantity of interest. As noted in \citep{henkes2022}, the scaling only affects the weighting of the loss terms and not the dynamics of the partial differential equations as long as it is linear. Once the model is trained, the outputs can be scaled back to their actual values. As for the machine learning software, we prepare codes from scratch using version $2.0$ of the TensorFlow library \citep{tensorflow}. 

\section{\label{2D}Illustrative problem and results}
\label{Sec:Results}
\begin{figure}
\centering
\includegraphics[width=0.8\textwidth,height=0.5\textheight,keepaspectratio]{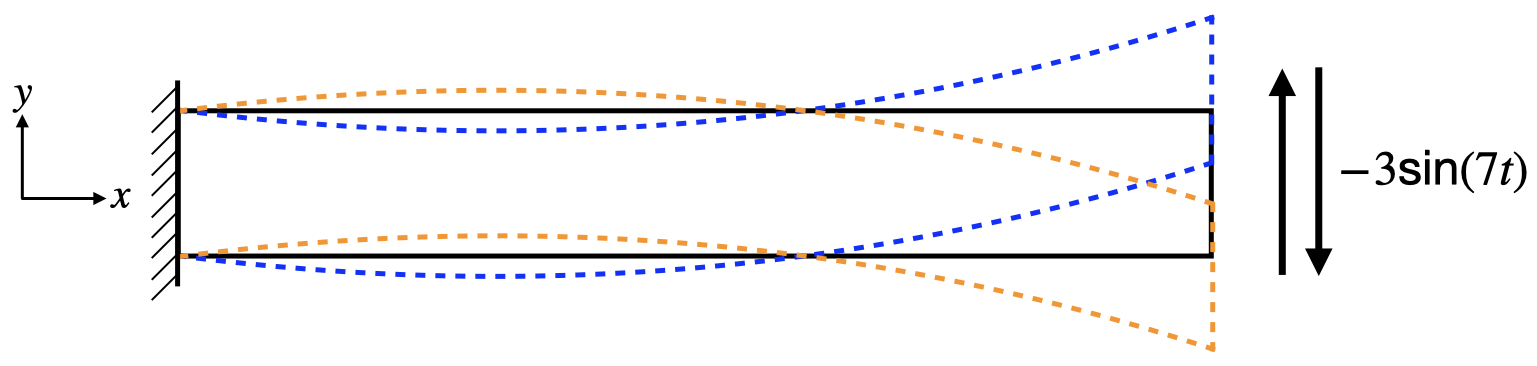}
\caption{\label{fig:schema1} \markup{Schematic of the two-dimensional (2D) cantilever beam under an oscillating excitation at the free end. For clarity, blue and orange lines show how the beam moves up and down. }}
\end{figure}
We consider a 2D cantilever beam of size 200 mm $\times$ 10 mm with $ x \in [0,200]  \ \text{mm}$, $y \in [0,10] \ \text{mm}$ subjected to oscillating excitation at the free end. It is shown as a schematic in the figure Fig. \ref{fig:schema1}. We assume a linearly elastic material with a density of $\rho=0.92 \times 10^{-6} \ \text{kg}/\text{mm}^3$. The material properties are assumed as $\lambda=0.533334$ MPa and $\mu=0.1$ MPa. For training the PINN model and benchmarking, we use FEM solution data from Ansys. Spatially, entire solution is computed on a grid with $51051$ nodes of which $2100$ nodes belong to the boundaries. The total simulation time is set for $3$ seconds. The units for displacements are in mm and stress are in MPa.  

\subsection{\label{2D_forward} PINN model for forward problem}

In this section, we will discuss the forward modeling part of the problem at hand. In practical industrial scenarios, one cannot avoid experiments and measurements of the manufactured components. Hence, we can assume to have prior information of the quantities of interest, for e.g. displacement time series albeit at a sparse set of sensor points on the material. Since we do not perform any experiments, for benchmarking our results from PINN, we use a small \% of FEM data only at the boundaries and compute the data loss along with the equation loss. Thus, one can think of the sparse FEM simulation data to be analogous to experimental data.

\begin{figure*}
\begin{subfigure}[t]{0.33\textwidth}
\includegraphics[width=\textwidth,height=\textheight,keepaspectratio]{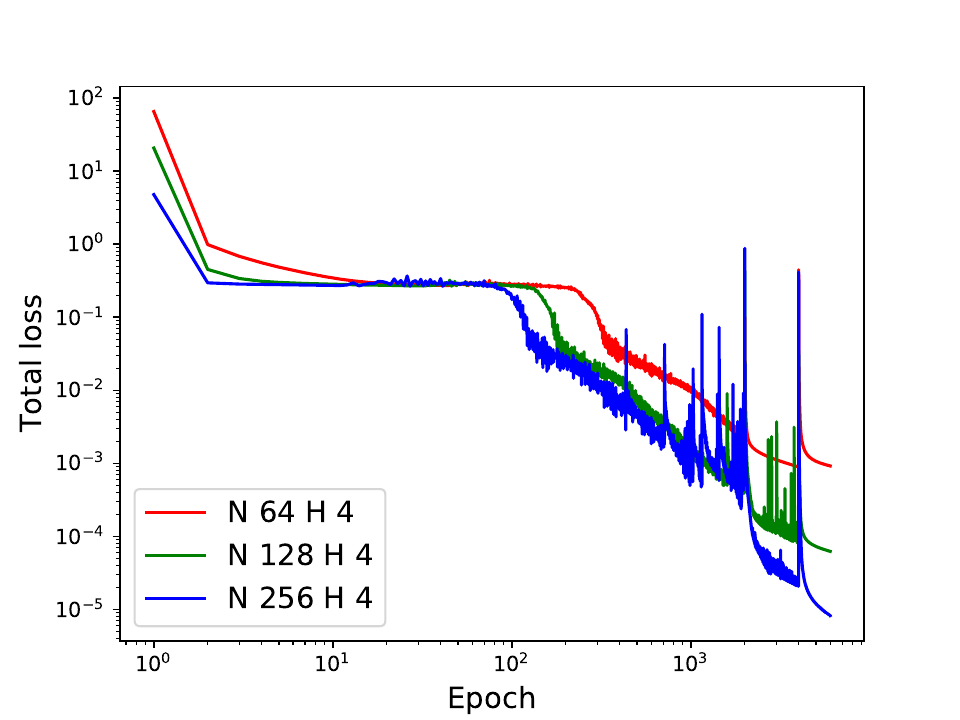}
\caption{}\label{D1_1}
\end{subfigure}
\begin{subfigure}[t]{0.33\textwidth}
\includegraphics[width=\textwidth,height=\textheight,keepaspectratio]{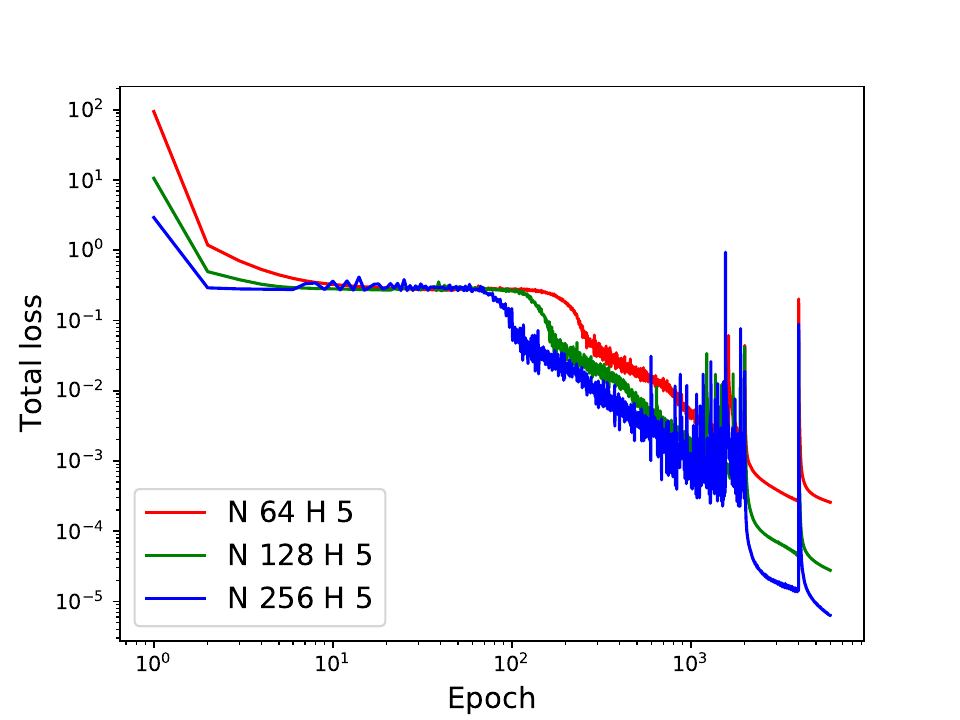}
\caption{}\label{D1_2}
\end{subfigure}
\begin{subfigure}[t]{0.33\textwidth}
\includegraphics[width=\textwidth,height=\textheight,keepaspectratio]{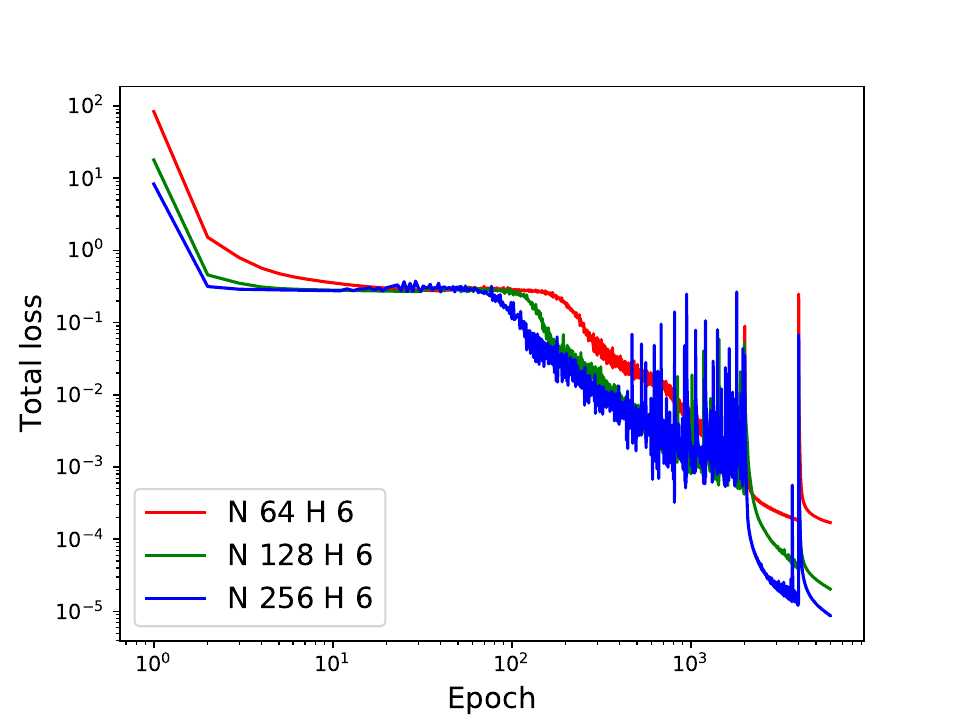}
\caption{}\label{D1_2}
\end{subfigure}
\caption{\label{fig:wide} {\markup{Total loss}} for different number of neurons and layers with $6.76$ \% training data.}
\label{fig:loss_hyperparam_d1}
\end{figure*}

\begin{figure*}
\begin{subfigure}[t]{0.33\textwidth}
\includegraphics[width=\textwidth,height=\textheight,keepaspectratio]{./D1_1.pdf}
\caption{}\label{D2_1}
\end{subfigure}
\begin{subfigure}[t]{0.33\textwidth}
\includegraphics[width=\textwidth,height=\textheight,keepaspectratio]{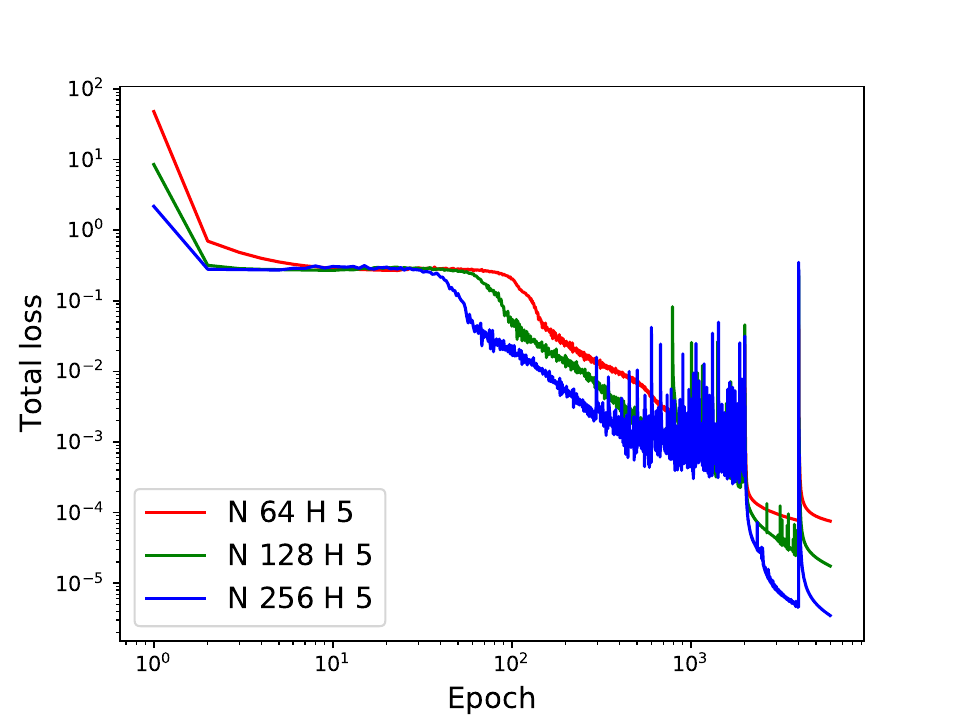}
\caption{}\label{D1_2}
\end{subfigure}
\begin{subfigure}[t]{0.33\textwidth}
\includegraphics[width=\textwidth,height=\textheight,keepaspectratio]{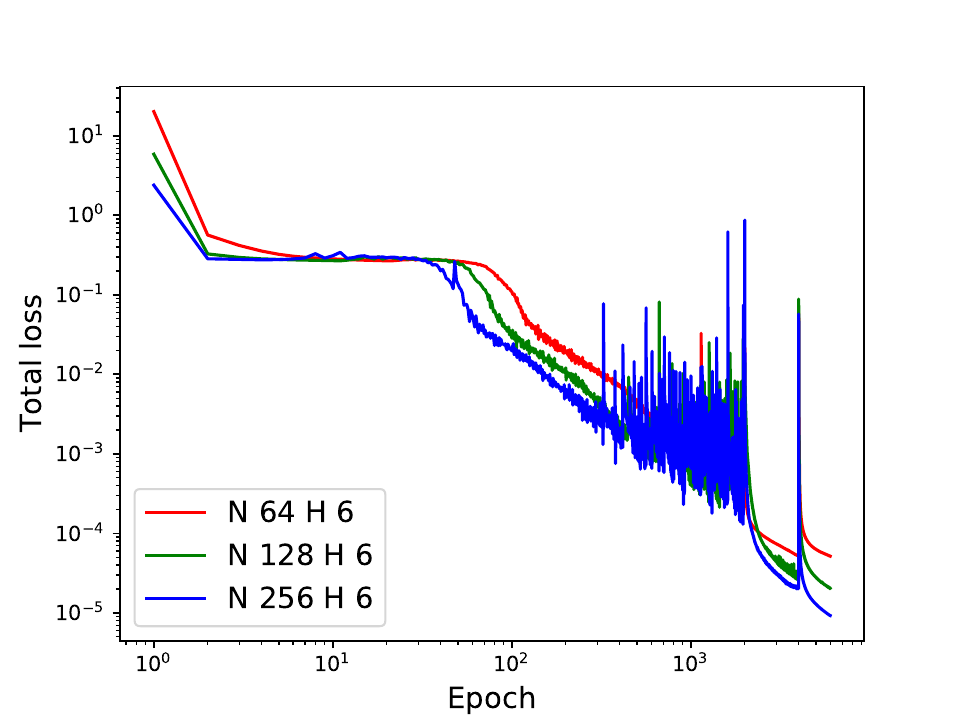}
\caption{}\label{D1_2}
\end{subfigure}
\caption{\label{fig:wide} {\markup{Total loss}} for different number of neurons and layers with $11.52$ \% training data.}
\label{fig:loss_hyperparam_d2}
\end{figure*}

\begin{figure*}
\begin{subfigure}[t]{0.33\textwidth}
\includegraphics[width=\textwidth,height=\textheight,keepaspectratio]{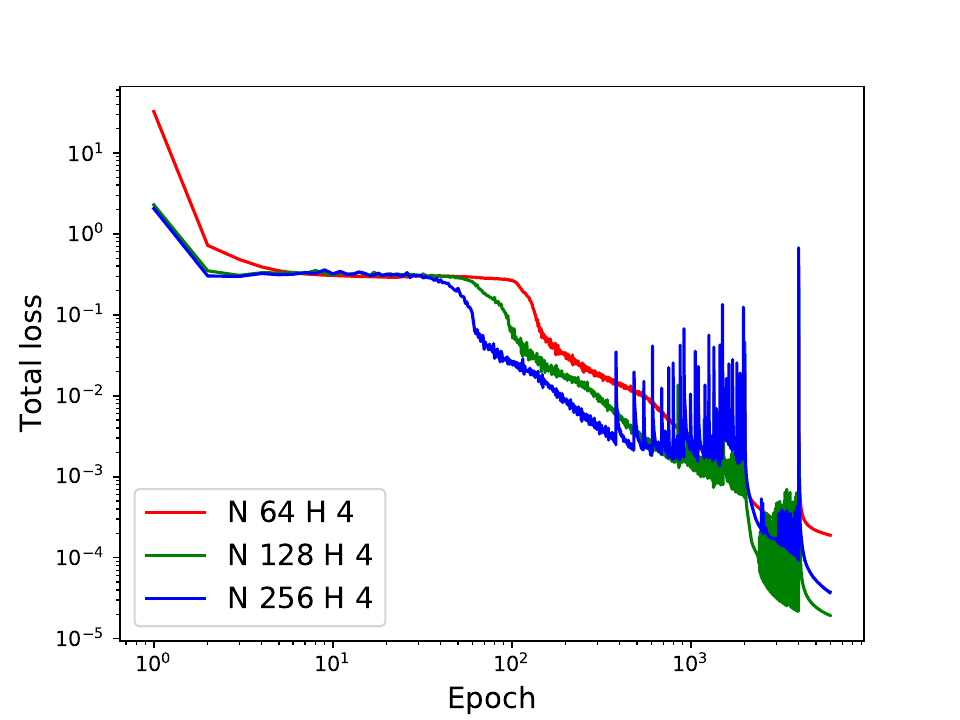}
\caption{}\label{D3_1}
\end{subfigure}
\begin{subfigure}[t]{0.33\textwidth}
\includegraphics[width=\textwidth,height=\textheight,keepaspectratio]{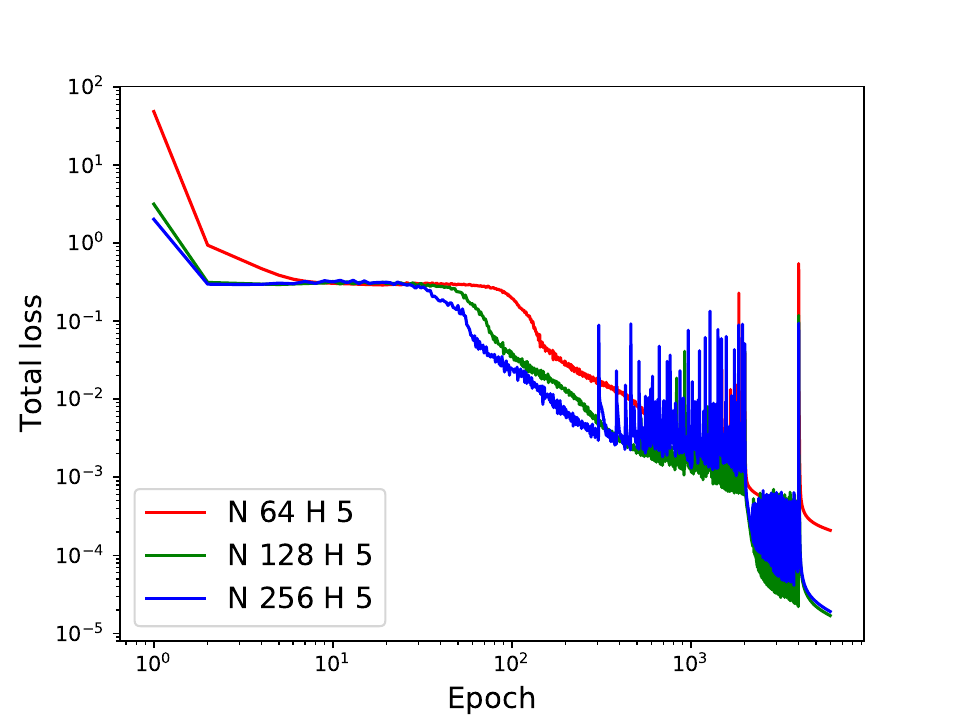}
\caption{}\label{D3_2}
\end{subfigure}
\begin{subfigure}[t]{0.33\textwidth}
\includegraphics[width=\textwidth,height=\textheight,keepaspectratio]{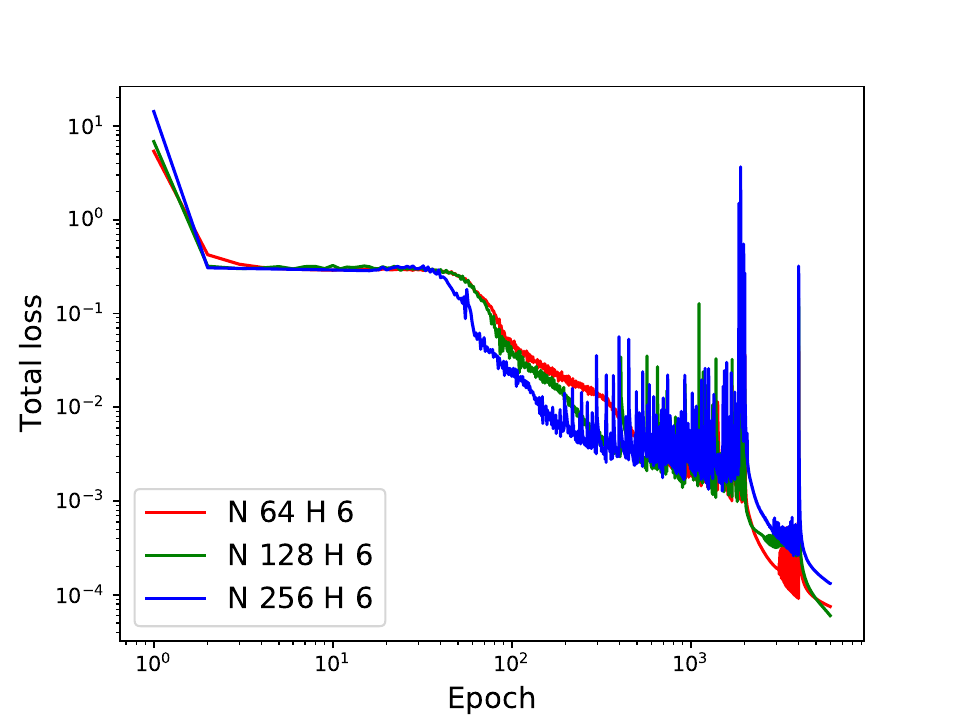}
\caption{}\label{D3_2}
\end{subfigure}
\caption{\label{fig:wide} {\markup{Total loss}} for different number of neurons and layers with $23.9$ \% training data.}
\label{fig:loss_hyperparam_d3}
\end{figure*}

\subsubsection{\label{2D_forward_1} Hyperparameter search}

For all our hyperparameter sweeps, we train the model for a total of $6000$ {\markup{epochs}}. For the first $2000$ {\markup{epochs}}, we use a learning rate of $\eta=0.001$. For the next $2000$ {\markup{epochs}}, we keep $\eta=0.0001$ and for the last $2000$ {\markup{epochs}}, $\eta=0.00001$. As for the activation functions we use a $\tanh$ function for all the layers except for the last layer where we use a linear activation function \citep{raissi19}. The training is performed using a NVIDIA Tesla V100 graphic card. We choose three sets of neurons $(64,128,256)$ per hidden layer and number of hidden layers $(4,5,6)$ for three training data sets. For the same, the total loss curves are shown in Figs. \ref{fig:loss_hyperparam_d1}, \ref{fig:loss_hyperparam_d2} and \ref{fig:loss_hyperparam_d3}. From Fig. \ref{fig:loss_hyperparam_d1}, for the training data of $6.76$ \% with batch size of 142, we find the computationally optimal set of neurons and layers to be $(256,4)$. Similarly, from Fig. \ref{fig:loss_hyperparam_d2}, for $11.52$ \% training data with batch size of 121, the best set of neurons and layers are $(128,4)$. Finally, from Fig. \ref{fig:loss_hyperparam_d3}, for $23.9$ \% training data with batch size of 251, it is $(128,5)$. Here the \% data is shown with respect to boundary data from the FEM solution. Also, we use $500$ random collocation points for the calculation of $\Losseqn$ at each gradient descent step. These collocation points are sampled using Latin hypercube sampling (LHS) method \citep{LHS}. 

\begin{table*}
\caption{\label{tab:table1}Best set of hyperparameters for different training data \% along with the corresponding errors when compared with the FEM solution.}
\begin{ruledtabular}
\begin{tabular}{lcccccc}
Num. neurons & Num. Layers & training data & collocation points  & $\epsilon_{u_x}$ & $\epsilon_{u_y}$ \\
\hline
$256$ & $4$ & $6.76$ \% & $500$ &  $3.12 \times 10^{-3}$ & $5.12 \times 10^{-4}$  \\
$128$ & $4$ & $11.56$ \% & $500$ &  $3.07 \times 10^{-3}$ & $5.48 \times 10^{-4}$ \\
$128$ & $5$ & $23.9$ \% & $500$ &  $2.91 \times 10^{-3}$ & $5.2 \times 10^{-4}$
\end{tabular}
\label{tab:hyperparameters_comp}
\end{ruledtabular}
\end{table*}

\subsubsection{\label{2D_forward_1} Comparison with FEM solution}

For assessing the accuracy of solutions based on PINN, we compare them with numerical solution based on FEM. From the hyperparameter search, we found three sets of neurons and layers which are optimal for three sets of training data.  For the same, we define an error metric known as the normalized root mean square error (NRMSE)
\begin{align}
    \epsilon = \frac{\biggl( \frac{1}{n_t} \sum_{i}^{n_t} (\bf{s}^{\text{PINN}}_{i} - \bf{s}^{\text{FEM}}_{i})^2 \biggr)^{1/2}}{\text{max}(\bf{s}^{\text{FEM}})-\text{min}(\bf{s}^{\text{FEM}}) },
    \label{eq:aRMSE}
\end{align}
where, $n_t$ is the total number of spatial or spatio-temporal points and $\bf{s}$ represents the quantity of interest. The reason we use this metric is because the numerical values of the quantities of interest vary by different orders of magnitude. 

For the best of hyperparameter sets for their corresponding training data \%, Table \ref{tab:hyperparameters_comp} shows the NRMSE error $\epsilon$ in $u_x$ and $u_y$ as compared with the FEM solution. We compute $\epsilon$ at 101 $\times$ 11 $\times$ 100 spatio-temporal points which corresponds to 101 $\times$ 11 sparse spatial grid points and 100 time instants. We observe that for all three training data sets, the testing space error is of the order $10^{-3}$ and $10^{-4}$ for the displacements $u_x$ and $u_y$ which is reasonably small. Errors for training data of $6.76$ \% is of the same order as that of other training data sets. So, for the following results in this section we have chosen the training data of $6.76$ \%. 

\begin{figure}
\begin{subfigure}[t]{0.33\textwidth}
\includegraphics[width=\textwidth,height=\textheight,keepaspectratio]{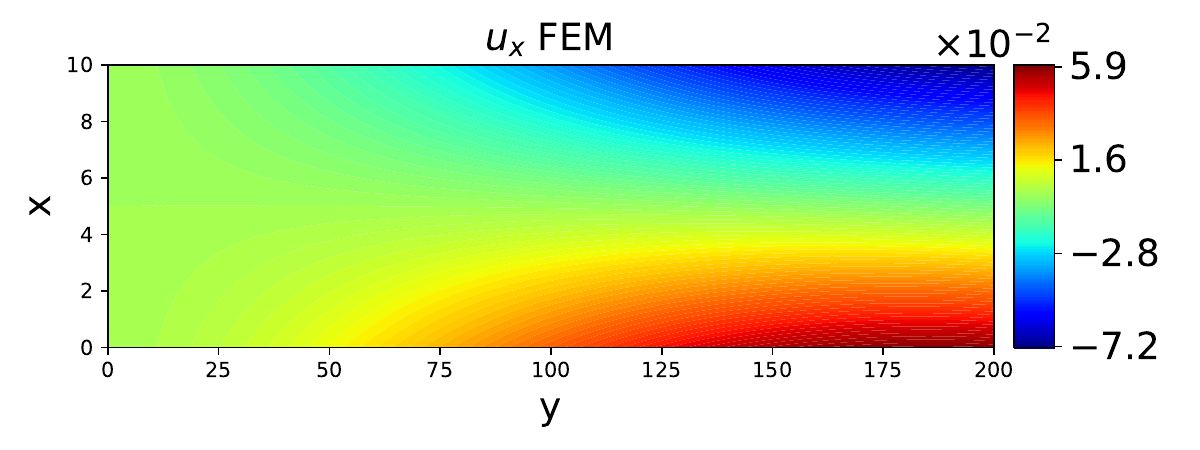}
\end{subfigure}
\begin{subfigure}[t]{0.33\textwidth}
\includegraphics[width=\textwidth,height=\textheight,keepaspectratio]{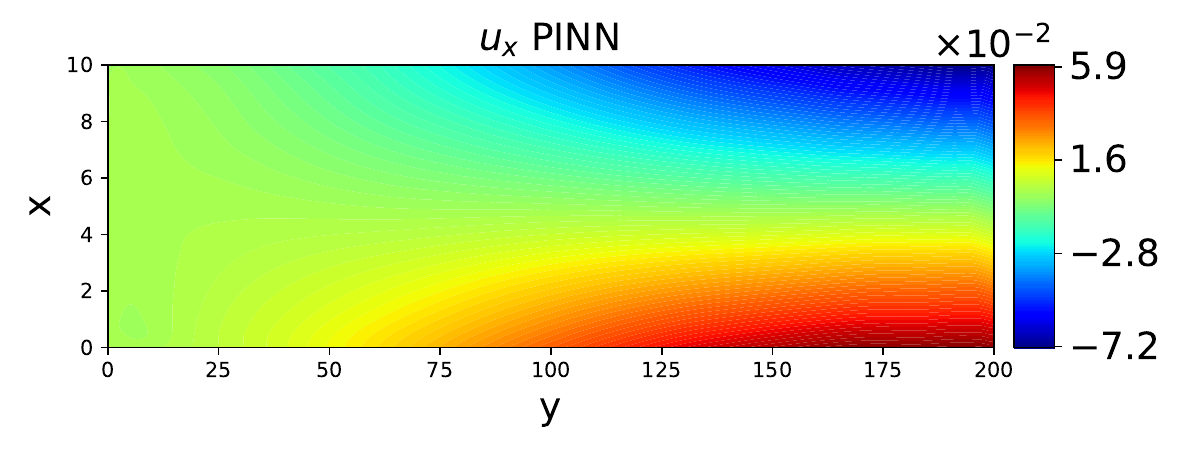}
\end{subfigure}
\begin{subfigure}[t]{0.33\textwidth}
\includegraphics[width=\textwidth,height=\textheight,keepaspectratio]{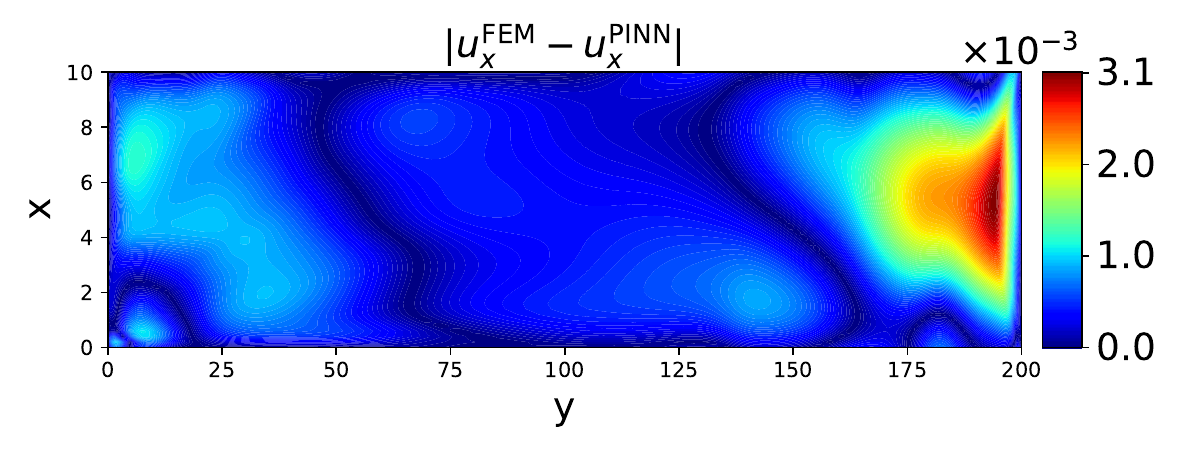}
\end{subfigure}
\begin{subfigure}[t]{0.33\textwidth}
\includegraphics[width=\textwidth,height=\textheight,keepaspectratio]{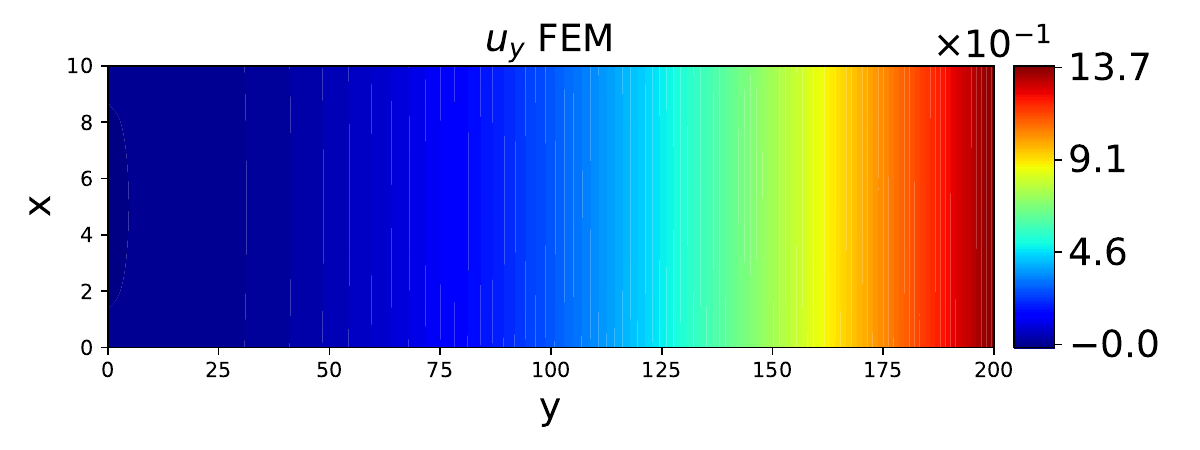}
\end{subfigure}
\begin{subfigure}[t]{0.33\textwidth}
\includegraphics[width=\textwidth,height=\textheight,keepaspectratio]{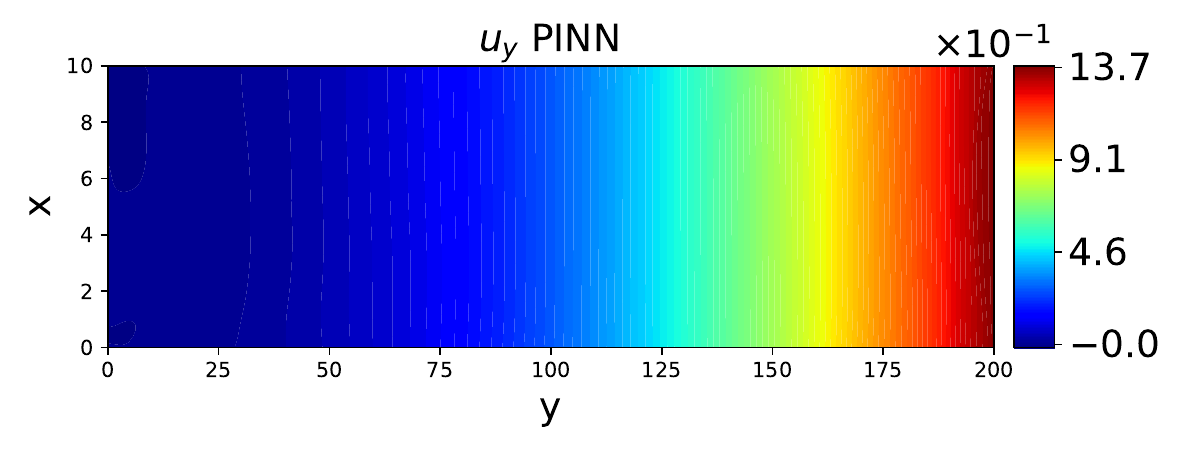}
\end{subfigure}
\begin{subfigure}[t]{0.33\textwidth}
\includegraphics[width=\textwidth,height=\textheight,keepaspectratio]{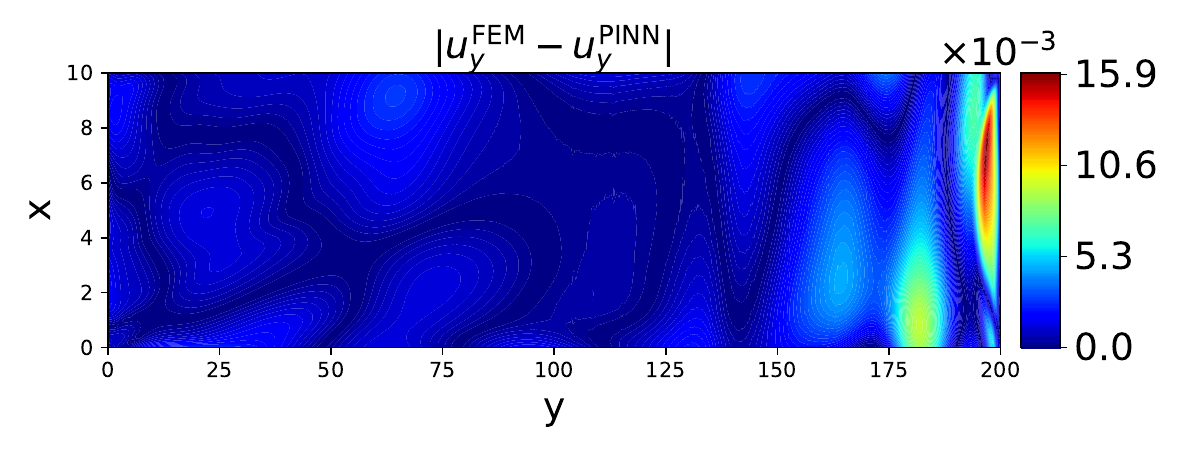}
\end{subfigure}
\begin{subfigure}[t]{0.33\textwidth}
\includegraphics[width=\textwidth,height=\textheight,keepaspectratio]{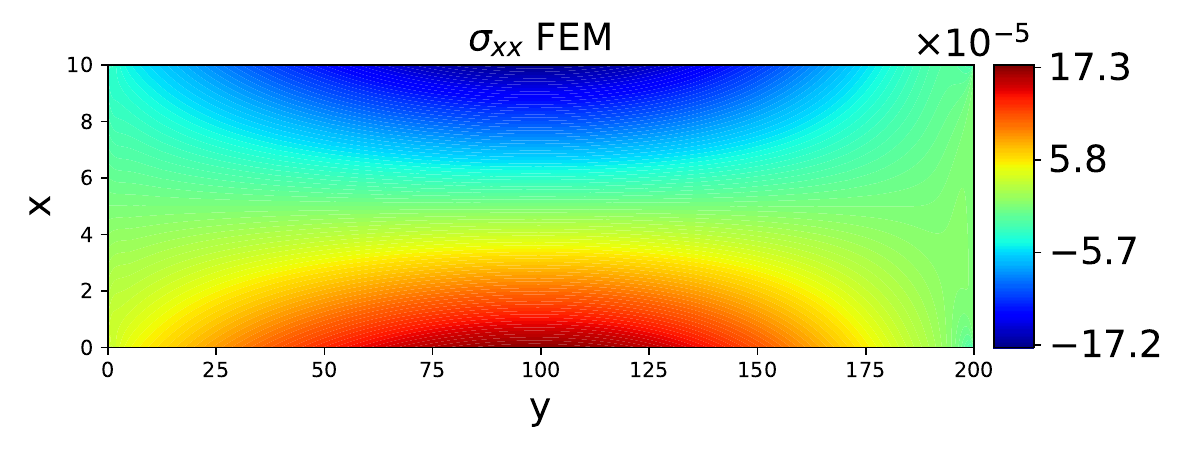}
\end{subfigure}
\begin{subfigure}[t]{0.33\textwidth}
\includegraphics[width=\textwidth,height=\textheight,keepaspectratio]{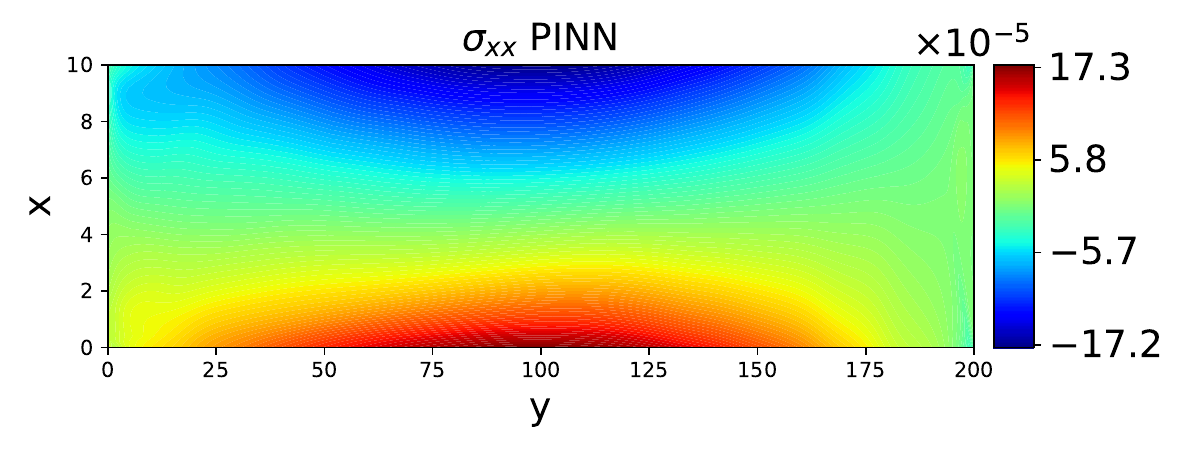}
\end{subfigure}
\begin{subfigure}[t]{0.33\textwidth}
\includegraphics[width=\textwidth,height=\textheight,keepaspectratio]{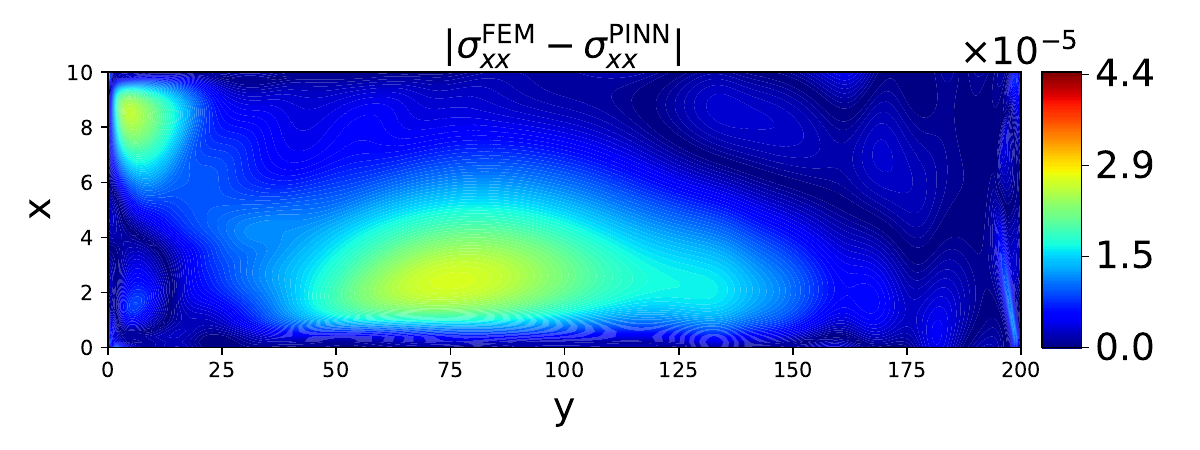}
\end{subfigure}
\begin{subfigure}[t]{0.33\textwidth}
\includegraphics[width=\textwidth,height=\textheight,keepaspectratio]{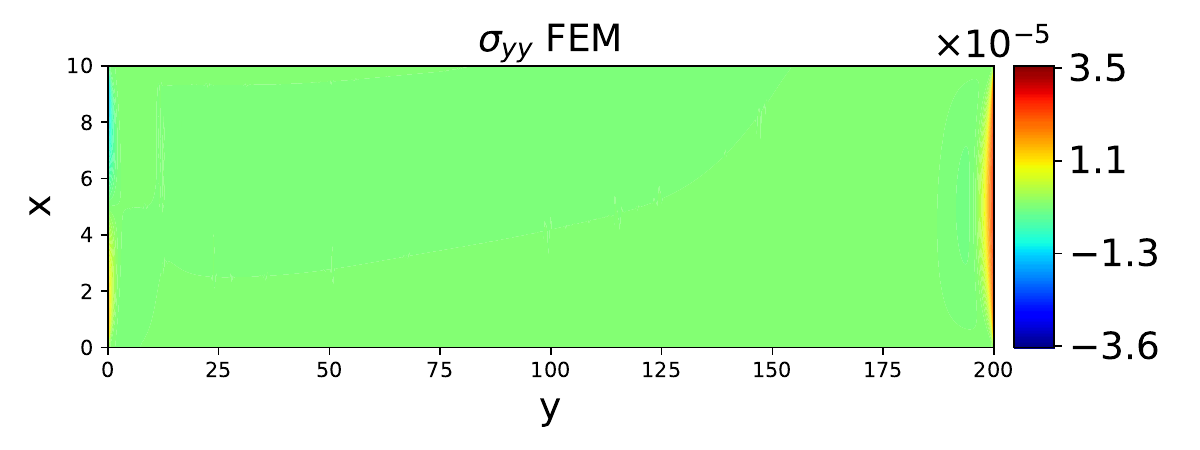}
\end{subfigure}
\begin{subfigure}[t]{0.33\textwidth}
\includegraphics[width=\textwidth,height=\textheight,keepaspectratio]{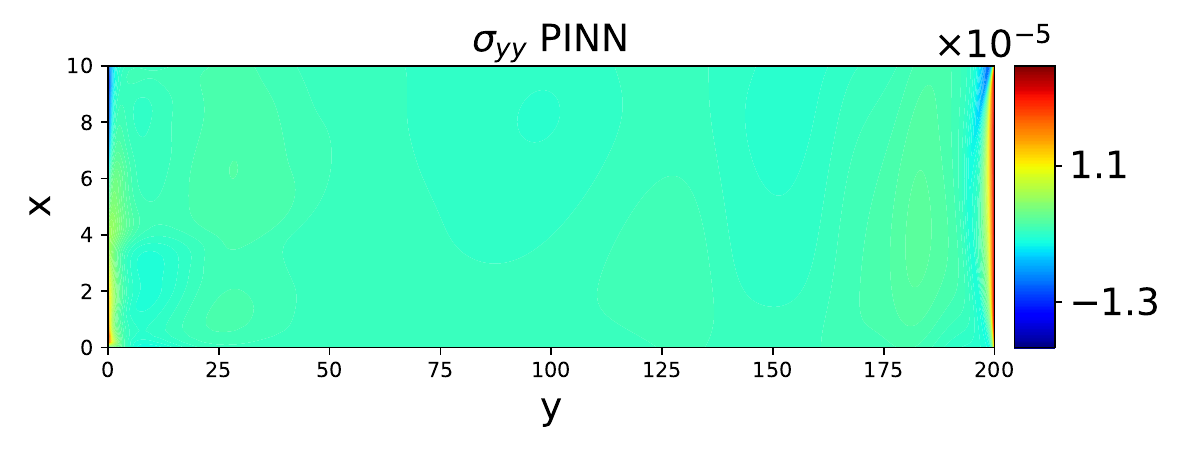}
\end{subfigure}
\begin{subfigure}[t]{0.33\textwidth}
\includegraphics[width=\textwidth,height=\textheight,keepaspectratio]{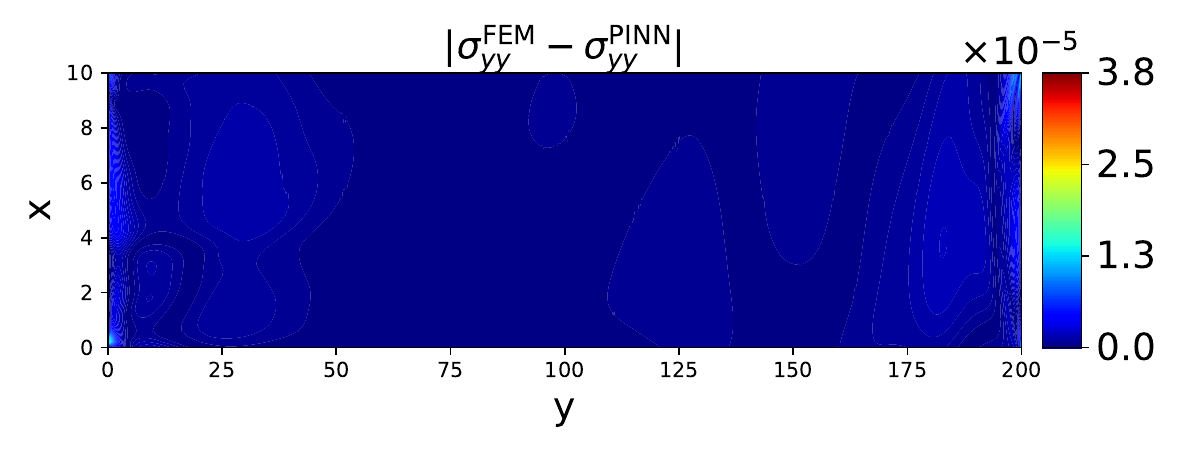}
\end{subfigure}
\begin{subfigure}[t]{0.33\textwidth}
\includegraphics[width=\textwidth,height=\textheight,keepaspectratio]{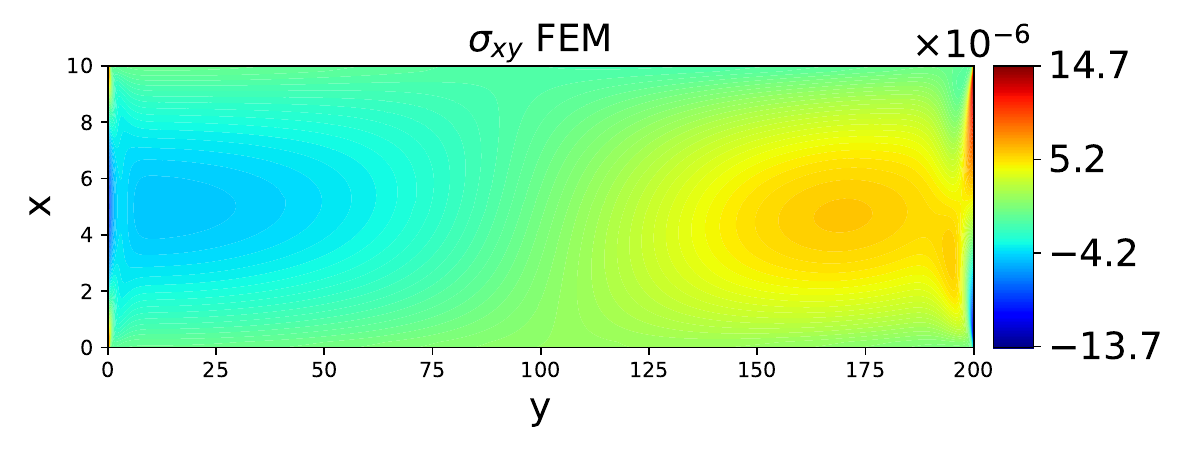}
\caption{}
\end{subfigure}
\begin{subfigure}[t]{0.33\textwidth}
\includegraphics[width=\textwidth,height=\textheight,keepaspectratio]{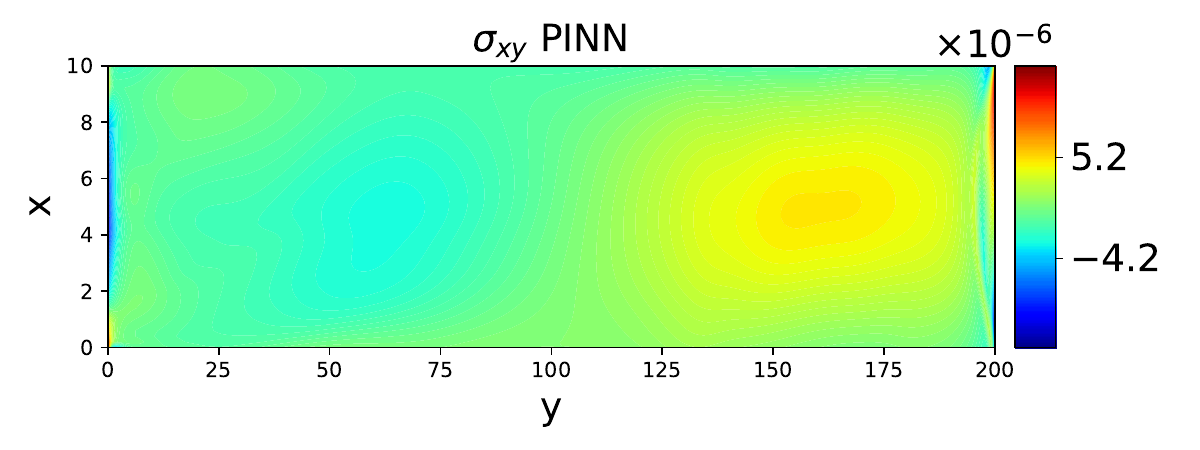}
\caption{}
\end{subfigure}
\begin{subfigure}[t]{0.33\textwidth}
\includegraphics[width=\textwidth,height=\textheight,keepaspectratio]{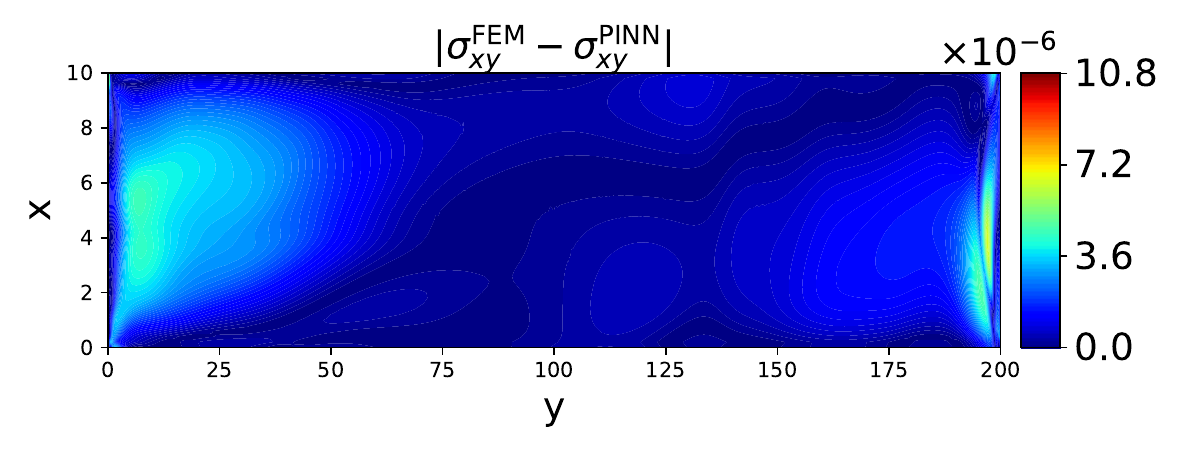}
\caption{}
\end{subfigure}
\caption{Contour plots of the quantities of interest at $t=1.72$ s for material properties $\lambda=0.533334$ MPa and $\mu=0.1$ MPa. Plots in (a) denote FEM solutions. Plots in (b) denote PINN solutions. Plots in (c) denotes the absolute error.}
\label{fig:contours_sec1}
\end{figure}

\begin{figure}
\begin{subfigure}[t]{0.49\textwidth}
\includegraphics[width=\textwidth,height=\textheight,keepaspectratio]{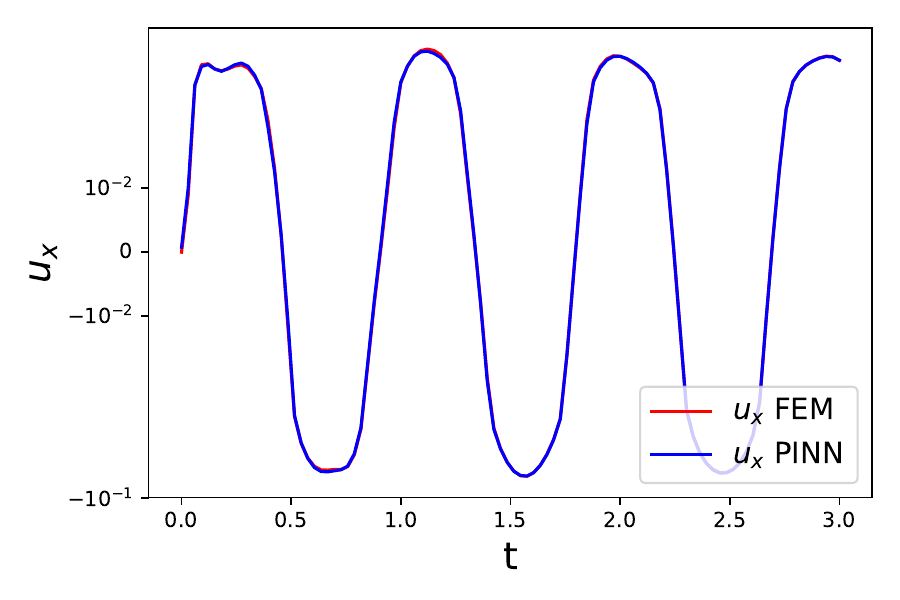}
\caption{}
\end{subfigure}
\begin{subfigure}[t]{0.49\textwidth}
\includegraphics[width=\textwidth,height=\textheight,keepaspectratio]{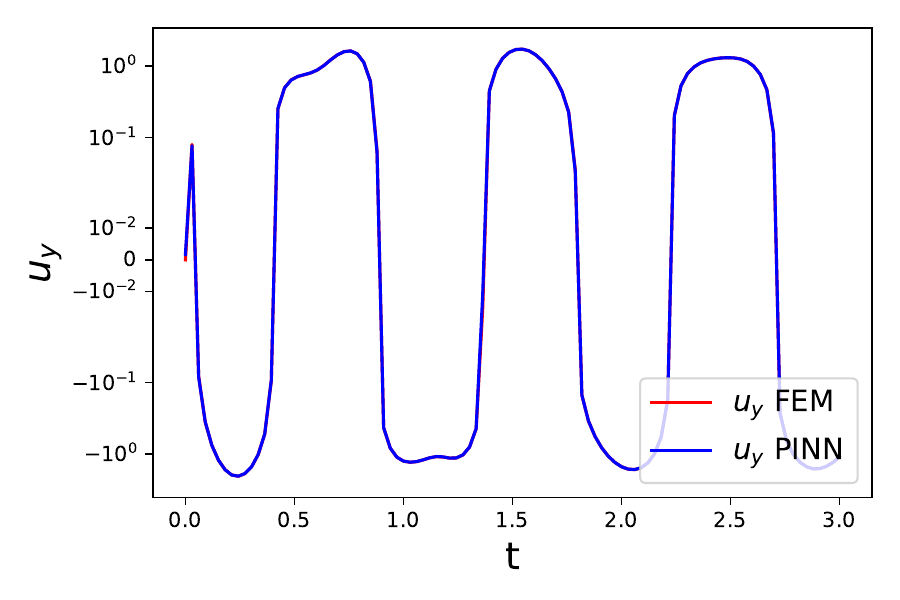}
\caption{}
\end{subfigure}
\begin{subfigure}[t]{0.33\textwidth}
\includegraphics[width=\textwidth,height=\textheight,keepaspectratio]{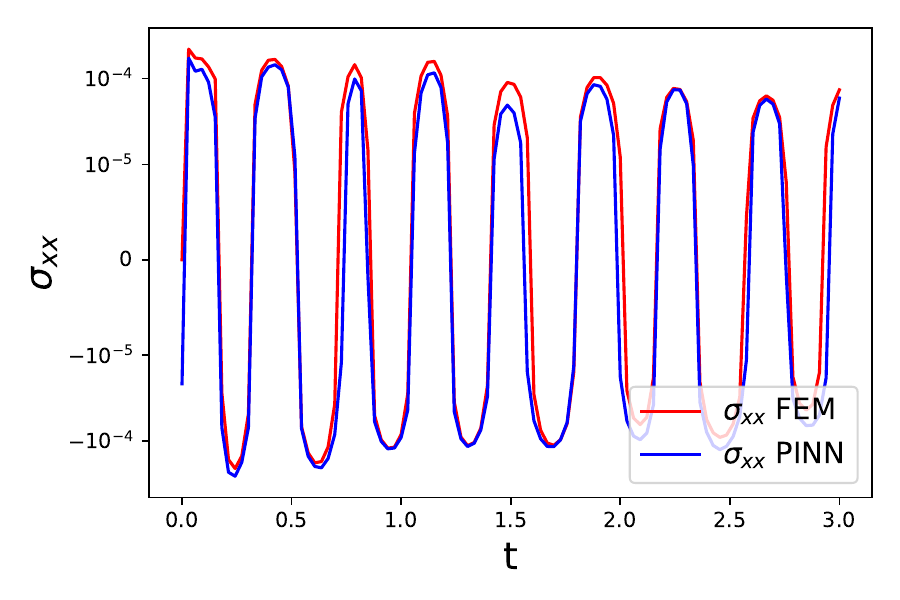}
\caption{}
\end{subfigure}
\begin{subfigure}[t]{0.33\textwidth}
\includegraphics[width=\textwidth,height=\textheight,keepaspectratio]{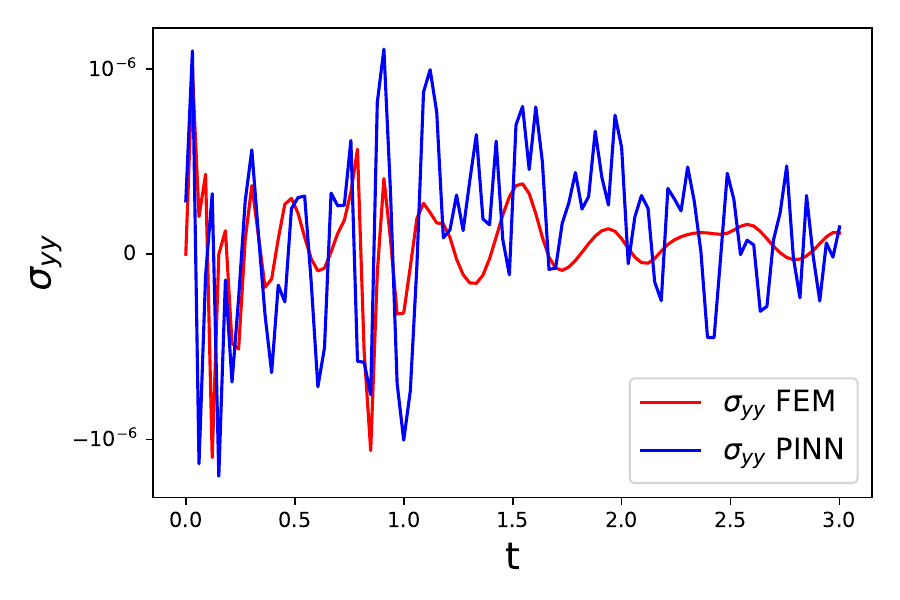}
\caption{}
\end{subfigure}
\begin{subfigure}[t]{0.33\textwidth}
\includegraphics[width=\textwidth,height=\textheight,keepaspectratio]{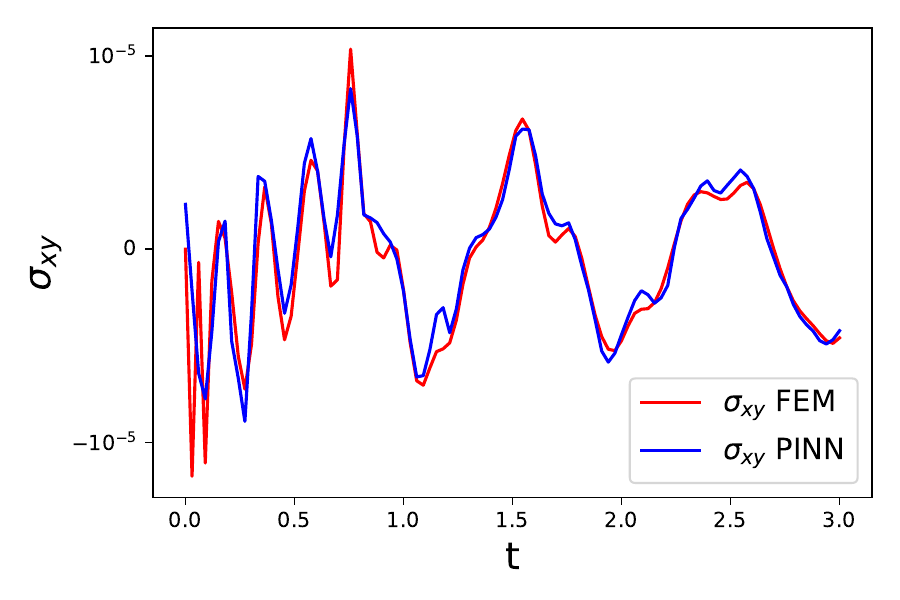}
\caption{}
\end{subfigure}
\caption{Line plots of the quantities of interest at $(x=120 \  \text{mm},\ y=8 \ \text{mm})$ for material properties $\lambda = 0.533334$ MPa and $\mu = 0.1$ MPa. Plots in (a) to (e) show the comparison of displacements and stresses obtained by PINN with respect to the FEM solutions.}
\label{fig:time_series_sec1}
\end{figure}

In Fig. \ref{fig:contours_sec1}, we compare the contour plots of displacements and stresses from the PINN model with the FEM solution at a particular time $t=1.51$ s. We observe an excellent agreement between PINN and FEM. Considering we used only a sparse set of training points on the boundary, it is remarkable that PINN is able to learn the solution at all the other interior spatial locations. Apart from the contour plots, we are also interested to see the time series of displacements and stresses at particular locations on the boundary of the beam. We show the same in Fig. \ref{fig:time_series_sec1}. We note that the displacements closely match with the FEM solution. Though the predicted stresses do not match the FEM solutions exactly, the PINN solutions are of the same order of magnitude. 

\begin{figure}
\begin{subfigure}[t]{0.49\textwidth}
\includegraphics[width=\textwidth,height=\textheight,keepaspectratio]{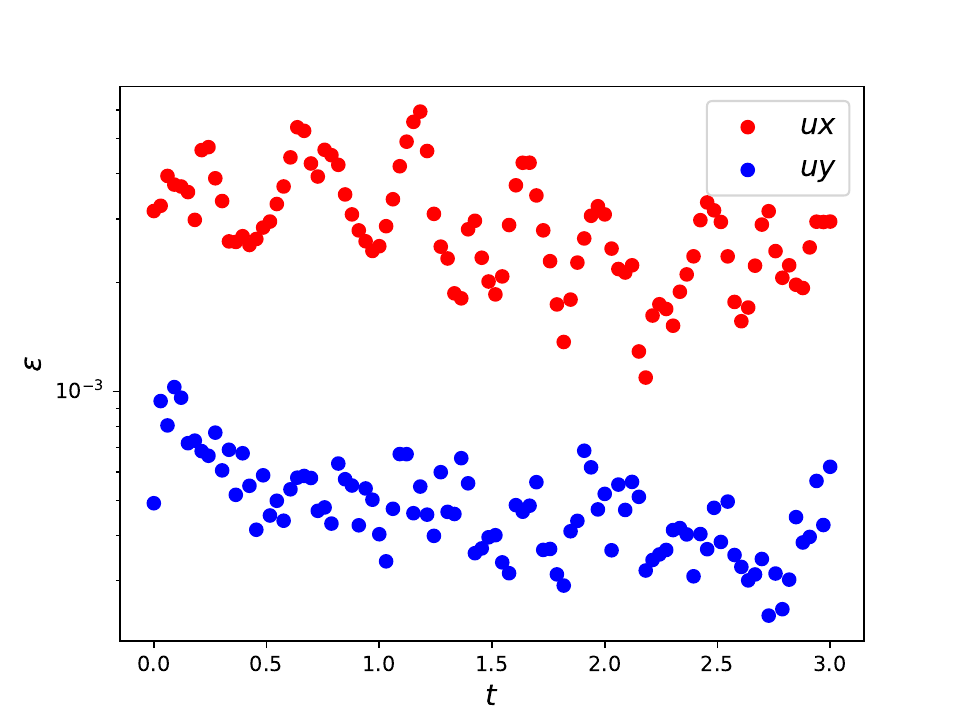}
\caption{}
\end{subfigure}
\begin{subfigure}[t]{0.49\textwidth}
\includegraphics[width=\textwidth,height=\textheight,keepaspectratio]{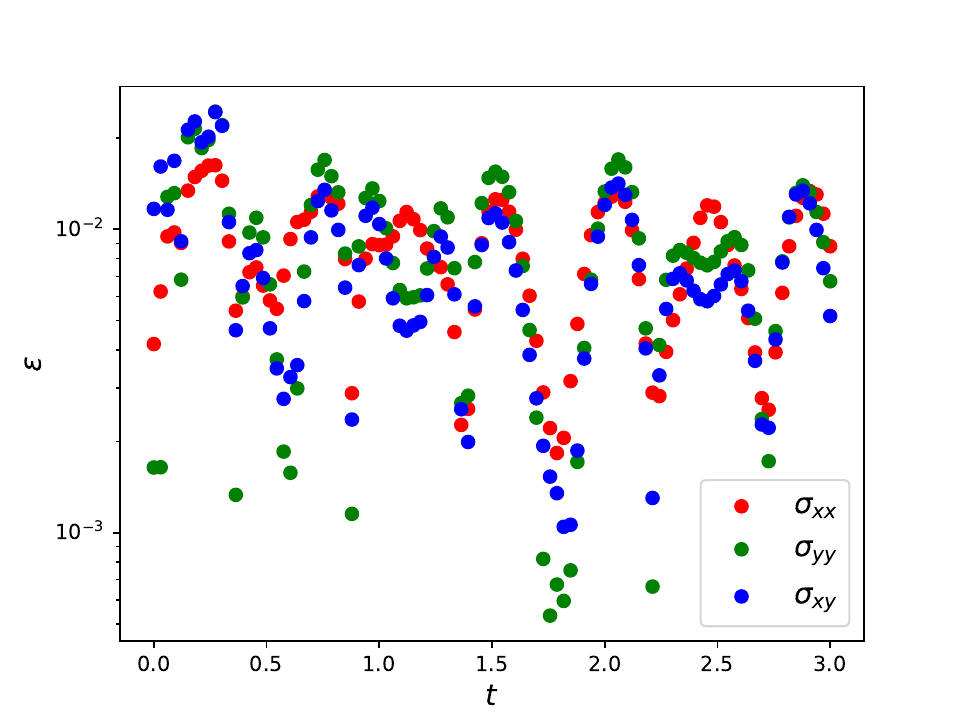}
\caption{}
\end{subfigure}
\caption{\label{fig:wide} Time series of errors for the material properties $\lambda=0.533334$ MPa and $\mu=0.1$ MPa. Plot (a) and (b) show the error in displacements and stresses respectively.}
\label{fig:nrmse_time_series_sec1}
\end{figure}

Next, we show the spatial NRMSE error $\epsilon$ in displacements and stresses with respect to time in Fig. \ref{fig:nrmse_time_series_sec1}. We observe that error is low and there is no general trend of error growth or decay in time. Considering we have shown convincing results from PINN for a sparse training data set of $6.76$ \%, if we analogously assume that we have a sparse set of measurement data, we could use this data along with a PINN model to obtain complete information of the displacements and stresses for all assumed spatio-temporal points for this dynamic system. This makes practical sense because we usually perform measurements at limited sensor locations due to cost considerations. Another advantage of the PINN model is that we get solution information at the interior of the structure which has similar accuracy to the FEM solution. 

\subsection{\label{2D_inverse} PINN model for inverse problem}
In this section, we discuss the results obtained using PINN in the context of inverse modeling where one knows the experimental data and wants to find the material properties ($\lambda, \mu$). Here we use FEM simulation data as an analogue of experimental data. The problem definition is same as shown in Fig. \ref{fig:schema1} except here we apply an oscillating excitation of $-\sin (2 \text{t})$. Also, in this case, we consider the material as steel which has a density of $7.85 \times 10^{-6}$ $kg/mm^{3}$ and it has the material properties $\lambda = 115385$ MPa and $\mu = 76923$ MPa. We run FEM simulation for $3.11$ seconds. For training the PINN model, we use all data from the boundaries as well as sparse data from interior of the domain which accounts for $4.4$ \% of entire simulation data. Also, temporally, we use the data at all the intermediate times which are available. As for the hyperparameters, we use $256$ neurons per hidden layer with $4$ hidden layers, and $1000$ random collocation points at each gradient descent step. Since we are interested in learning the material parameters, we consider these as trainable parameters along with weights and biases of the neural network. We consider $N_{\lambda}$, $N_{\mu}$ as trainable parameters where $\lambda^* = f(N_{\lambda}), \mu^* = f(N_{\mu})$. The choice of function $f$ significantly affects the learning process. Once the model is trained we can rescale these parameters to get actual material parameters by $\lambda = \lambda_c \lambda^*$ and $\mu = \mu_c \mu^*$.

As for the training configurations, we consider $N_{u_x}, N_{u_y}$ as neural network outputs for displacement in $x$ and $y$. We assume function $f$ to be linear and nonlinear (sigmoid and tanh). Coupled with this, we also tinkered with the boundary conditions by enforcing Dirichlet-type condition via hard constraints on displacements \citep{micro} like $u_x^* = x N_{u_x}$, $u_y^* = x N_{u_y}$ as the beam is fixed at one end resulting $u_x(0,y,t)=u_y(0,y,t)=0$. Also, we consider a case where there are no constrains on displacement outputs $u_x^* = N_{u_x}$, $u_y^* = N_{u_y}$. We assume an arbitrary scaling constant $\lambda_c = 150000$ MPa. 

We trained the PINN model for all combinations of $f$ and the boundary condition treatments as shown in Table \ref{tab:inverse_configs}. We find only the combination of $f=\text{sigmoid}$ with enforced hard boundary conditions gives accurate convergence of the material parameters while rest of the combinations failed. The reason is that the model converges to a local minima that satisfies the governing equations but are not realistic and it leads to a condition where parameters follow  $\lambda + 2\mu \sim 0 $ and hence parameters are not learned properly. We can solve this problem by putting constraints in learning parameters by using positive valued functions like sigmoid.

\begin{table*}
\caption{\label{tab:table1} Combinations of the functions $f$ and boundary condition treatments.}
\begin{ruledtabular}
\begin{tabular}{lccc}
 & linear & sigmoid & tanh \\
\hline
Hard constraints on BCs & \xmark & \cmark & \xmark \\
No hard constraints on BCs & \xmark & \xmark & \xmark \\
\end{tabular}
\label{tab:inverse_configs}
\end{ruledtabular}
\end{table*}

\begin{figure}
\begin{subfigure}[t]{0.33\textwidth}
\includegraphics[width=\textwidth,height=\textheight,keepaspectratio]{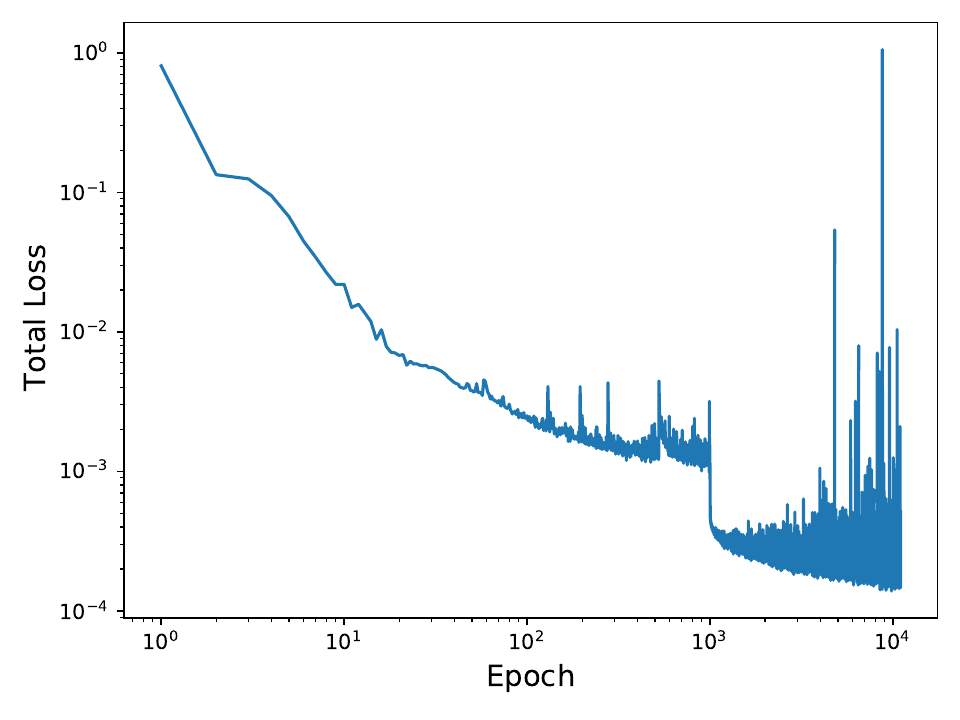}
\caption{}
\end{subfigure}
\begin{subfigure}[t]{0.33\textwidth}
\includegraphics[width=\textwidth,height=\textheight,keepaspectratio]{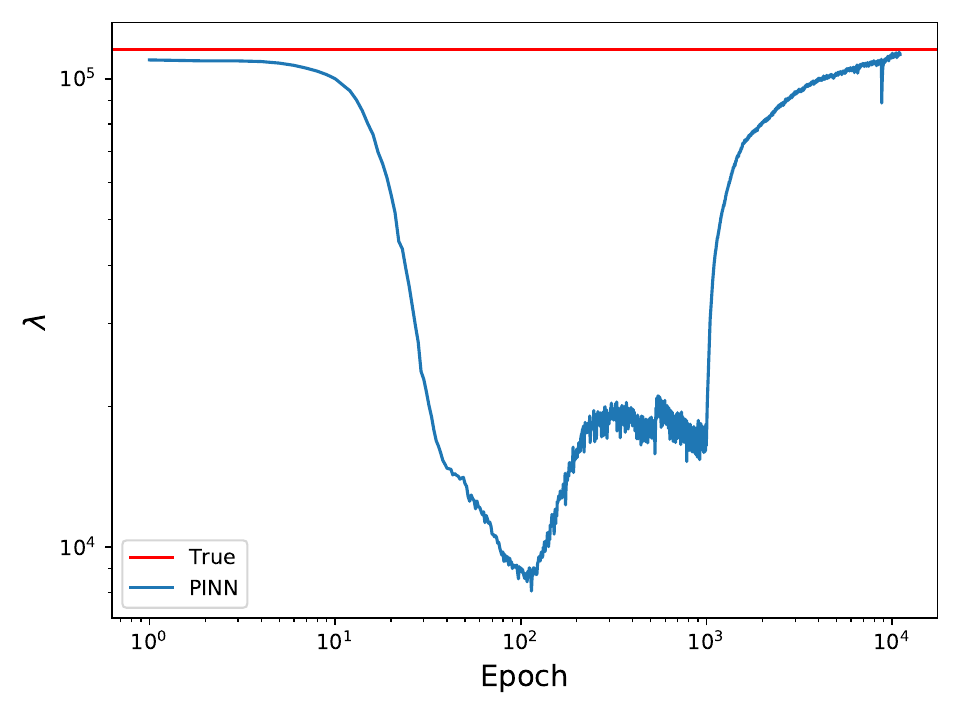}
\caption{}
\end{subfigure}
\begin{subfigure}[t]{0.33\textwidth}
\includegraphics[width=\textwidth,height=\textheight,keepaspectratio]{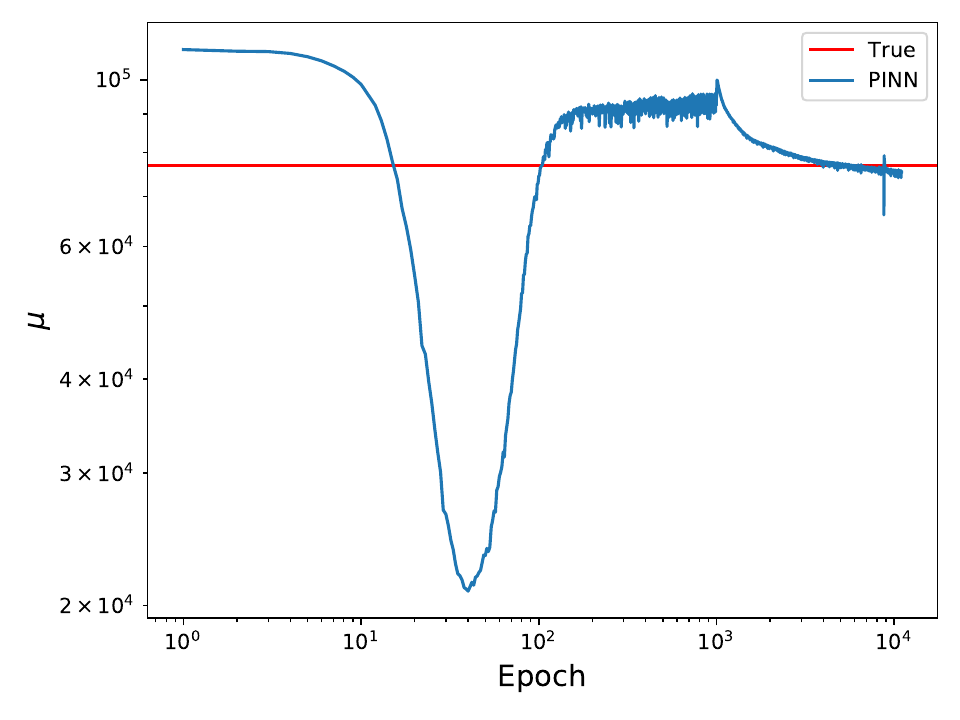}
\caption{}
\end{subfigure}

\caption{\label{fig:wide} Plot (a) shows the total loss, plots (b) and (c) show the evolution of predicted material properties $\lambda$ and $\mu$ over the course of training epochs.}
\label{fig:inverse_param_plot}
\end{figure}

We train the PINN model for first 1000 epochs with learning rate $\eta = 0.001$ and next 10000 epochs with $\eta = 0.0001$. The total training data is 4.4 \% of entire simulation data. We have considered batch size of 225 for the training.
We show in Fig. \ref{fig:inverse_param_plot} the total loss and the evolution of the predicted parameters $\lambda$ and $\mu$ over the course of training epochs. We observe that as the training progresses, the parameters tend to reach closer to their true value which is shown as a horizontal line. We get the PINN prediction error for $\lambda$ and $\mu$ to be $2.57$ \% and $1.9$ \% respectively. Although we find the prospect of PINN model's use in inverse modeling exciting, in this particular application, we find its practical use to be limited in the industrial context. One reason for this is that in most industrial settings, stress measurements inside the material which is required for training is not straightforward, especially if the geometry is complicated. Secondly, it can result in increased experimental costs because as more data is needed, more sensors are required. Due to these reasons, data quantity and quality for training the PINN model in this context will be limited. Hence, we will not expand upon the results of inverse models further in this study. However, the results we obtained in this context could be exciting for applications based on tomography. 

\subsection{\label{2D_surrogate} PINN based surrogate model for material property estimation}
\begin{figure}
\begin{subfigure}[t]{0.33\textwidth}
\includegraphics[width=\textwidth,height=\textheight,keepaspectratio]{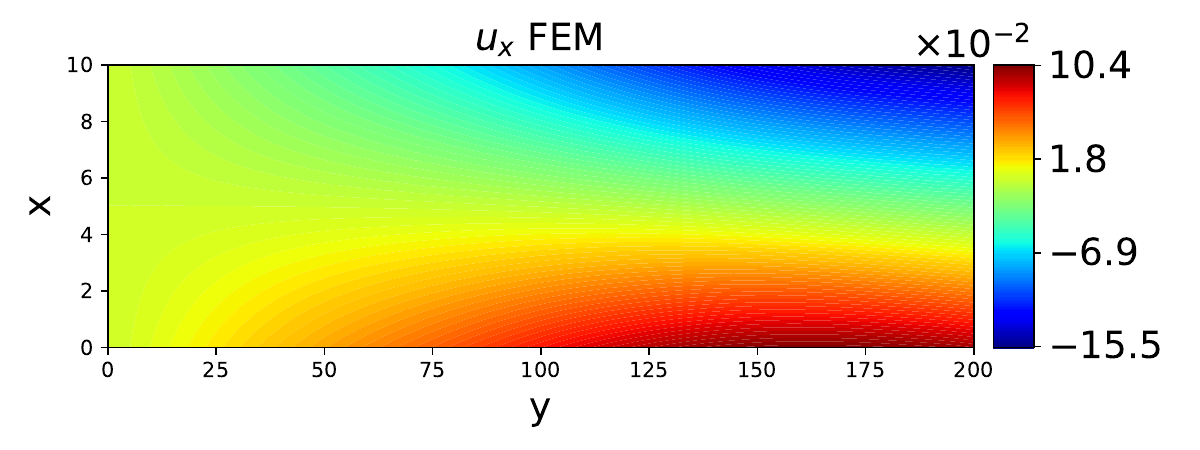}
\end{subfigure}
\begin{subfigure}[t]{0.33\textwidth}
\includegraphics[width=\textwidth,height=\textheight,keepaspectratio]{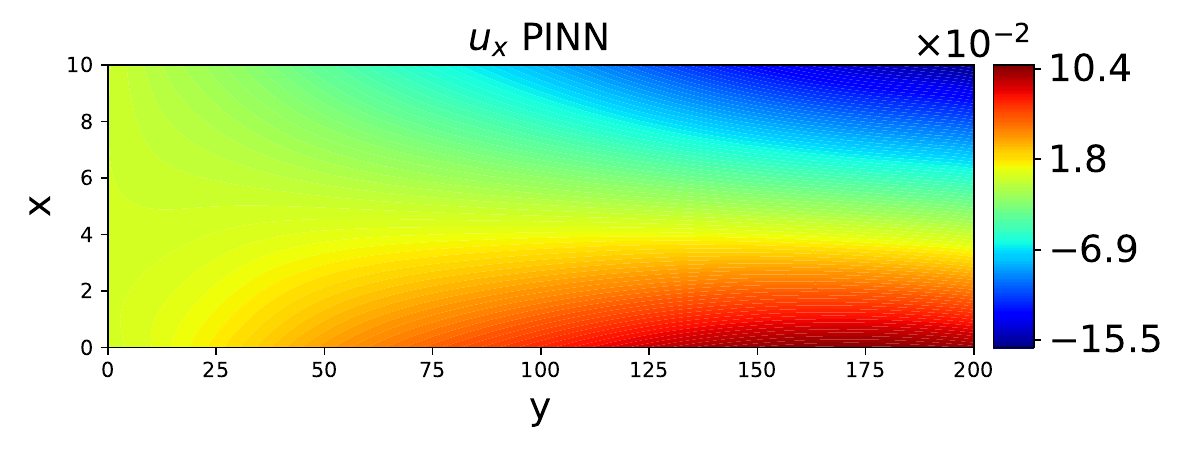}
\end{subfigure}
\begin{subfigure}[t]{0.33\textwidth}
\includegraphics[width=\textwidth,height=\textheight,keepaspectratio]{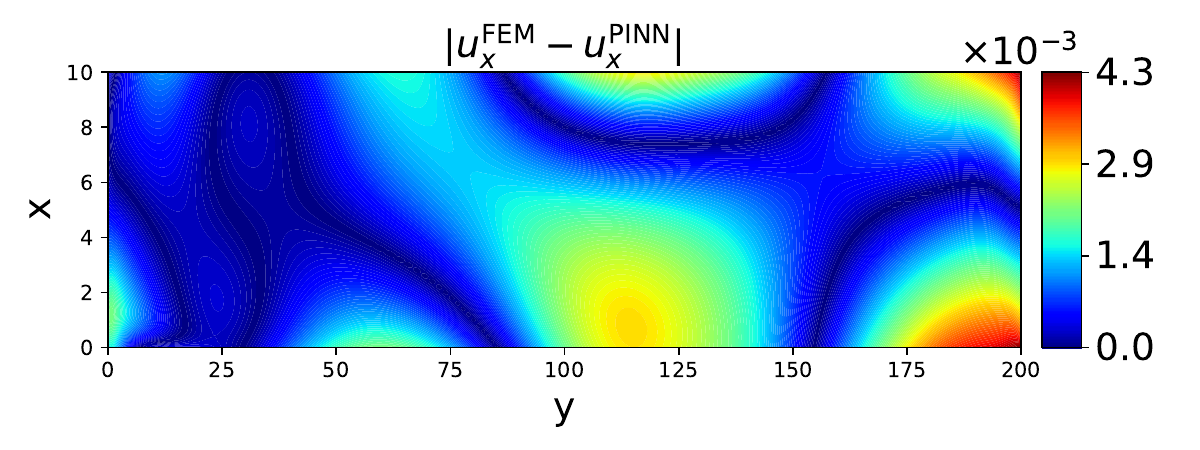}
\end{subfigure}
\begin{subfigure}[t]{0.33\textwidth}
\includegraphics[width=\textwidth,height=\textheight,keepaspectratio]{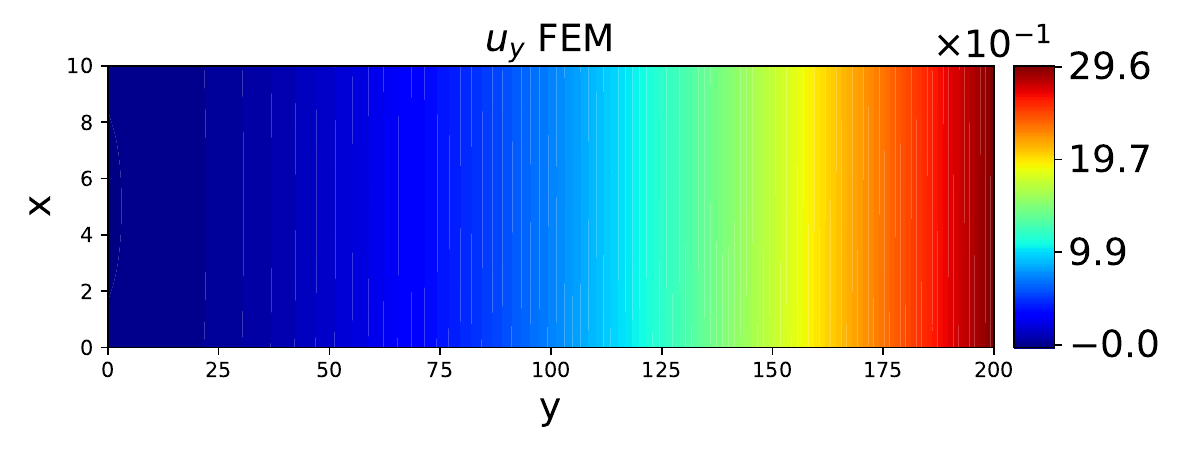}
\end{subfigure}
\begin{subfigure}[t]{0.33\textwidth}
\includegraphics[width=\textwidth,height=\textheight,keepaspectratio]{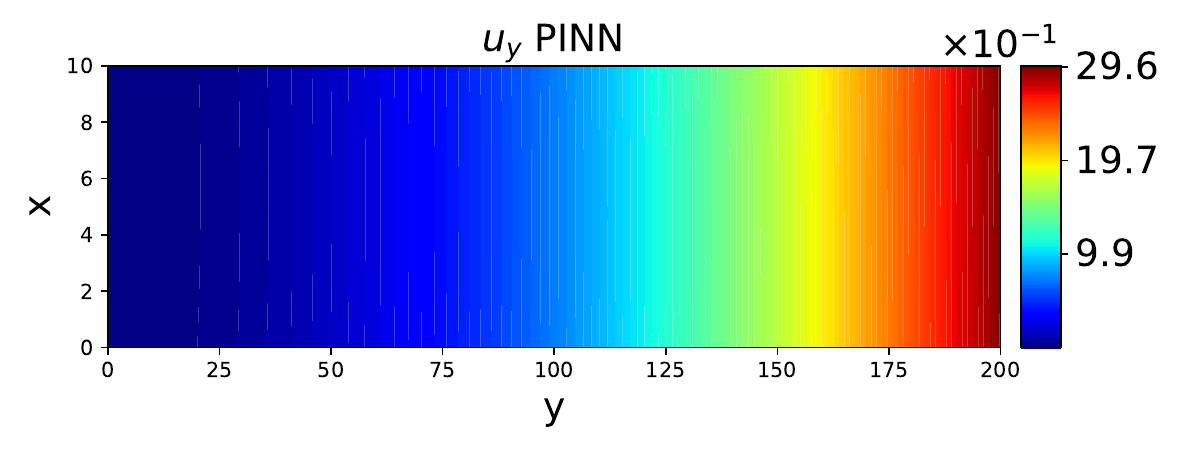}
\end{subfigure}
\begin{subfigure}[t]{0.33\textwidth}
\includegraphics[width=\textwidth,height=\textheight,keepaspectratio]{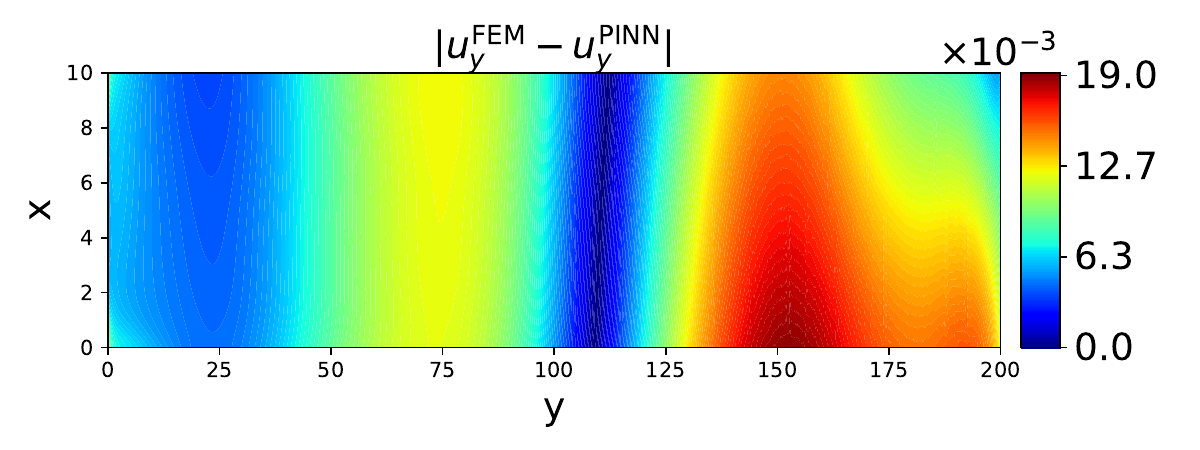}
\end{subfigure}
\begin{subfigure}[t]{0.33\textwidth}
\includegraphics[width=\textwidth,height=\textheight,keepaspectratio]{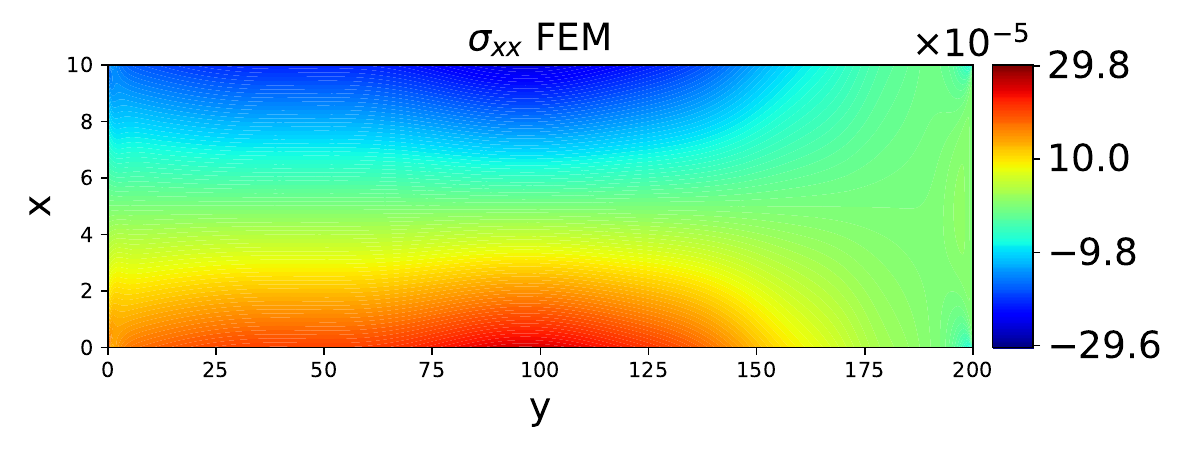}
\end{subfigure}
\begin{subfigure}[t]{0.33\textwidth}
\includegraphics[width=\textwidth,height=\textheight,keepaspectratio]{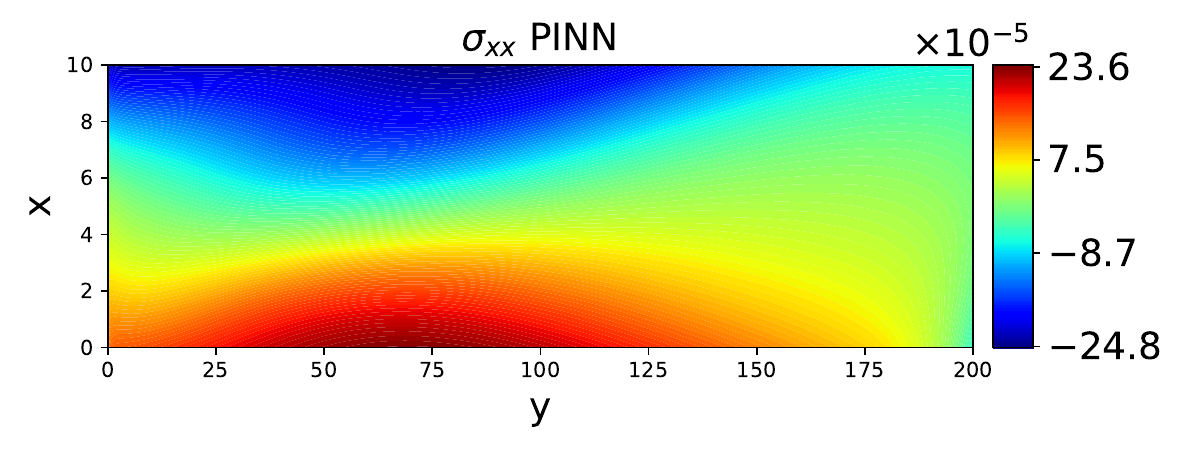}
\end{subfigure}
\begin{subfigure}[t]{0.33\textwidth}
\includegraphics[width=\textwidth,height=\textheight,keepaspectratio]{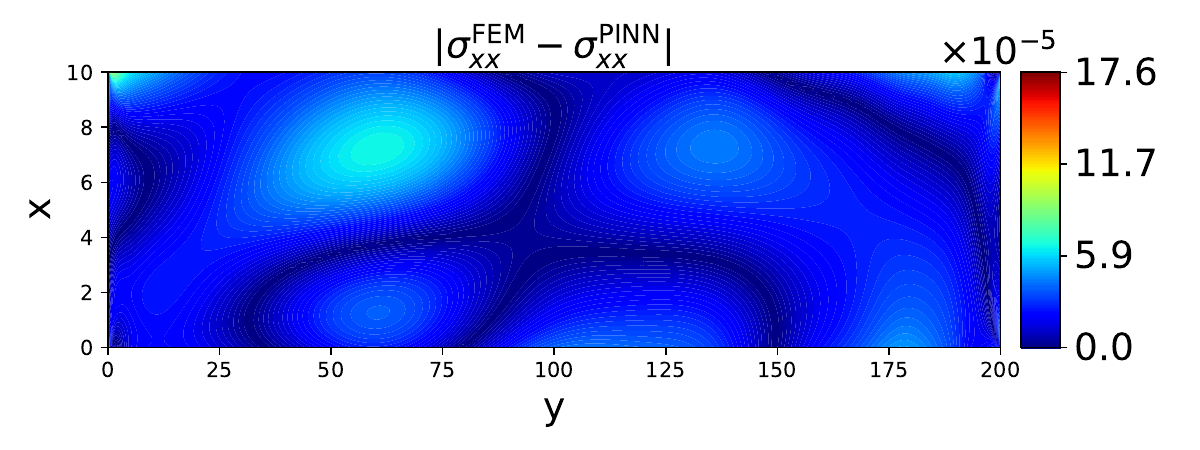}
\end{subfigure}
\begin{subfigure}[t]{0.33\textwidth}
\includegraphics[width=\textwidth,height=\textheight,keepaspectratio]{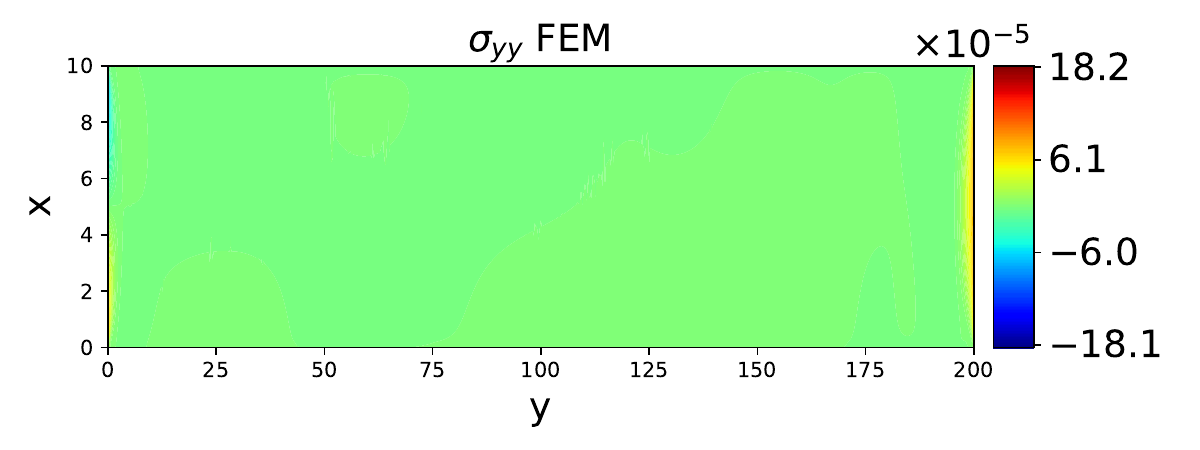}
\end{subfigure}
\begin{subfigure}[t]{0.33\textwidth}
\includegraphics[width=\textwidth,height=\textheight,keepaspectratio]{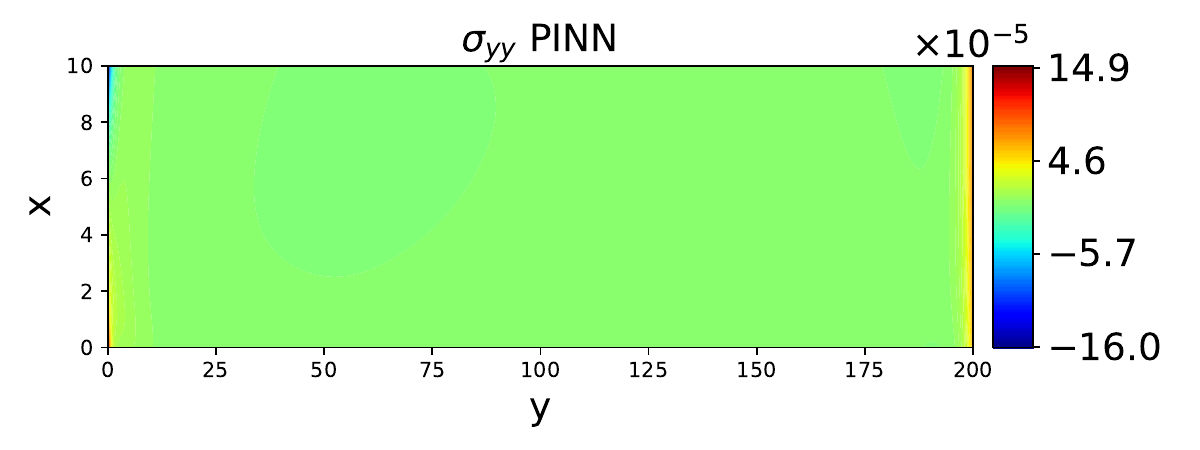}
\end{subfigure}
\begin{subfigure}[t]{0.33\textwidth}
\includegraphics[width=\textwidth,height=\textheight,keepaspectratio]{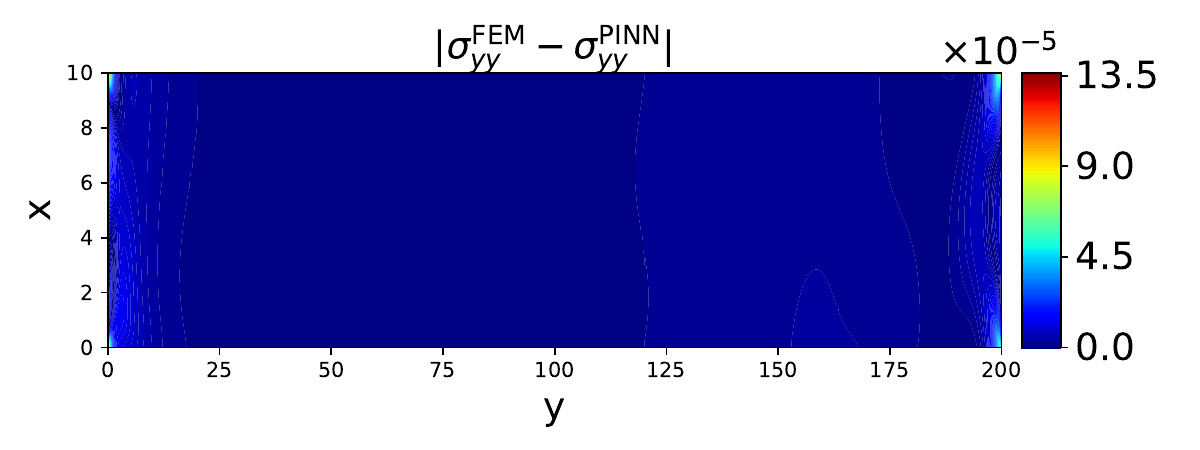}
\end{subfigure}
\begin{subfigure}[t]{0.33\textwidth}
\includegraphics[width=\textwidth,height=\textheight,keepaspectratio]{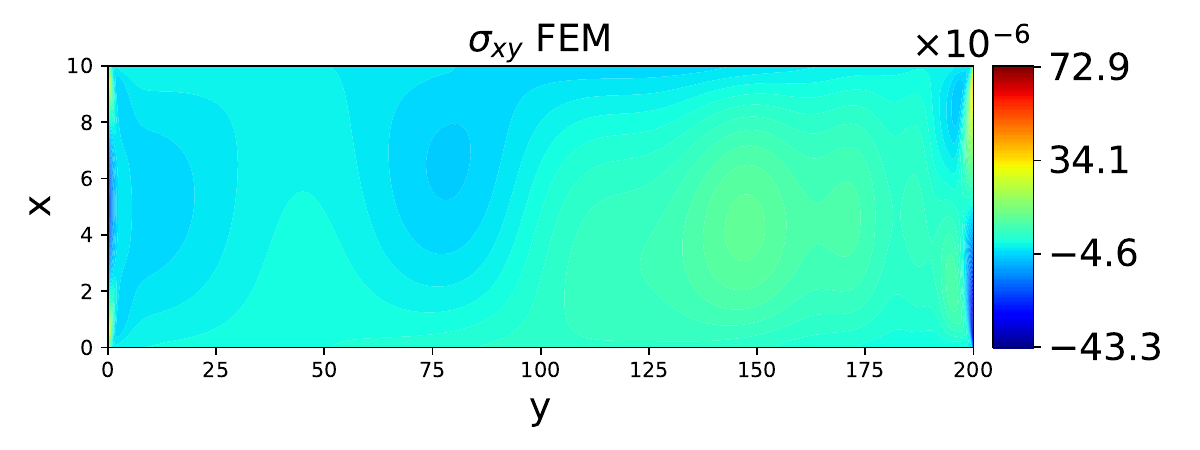}
\caption{}
\end{subfigure}
\begin{subfigure}[t]{0.33\textwidth}
\includegraphics[width=\textwidth,height=\textheight,keepaspectratio]{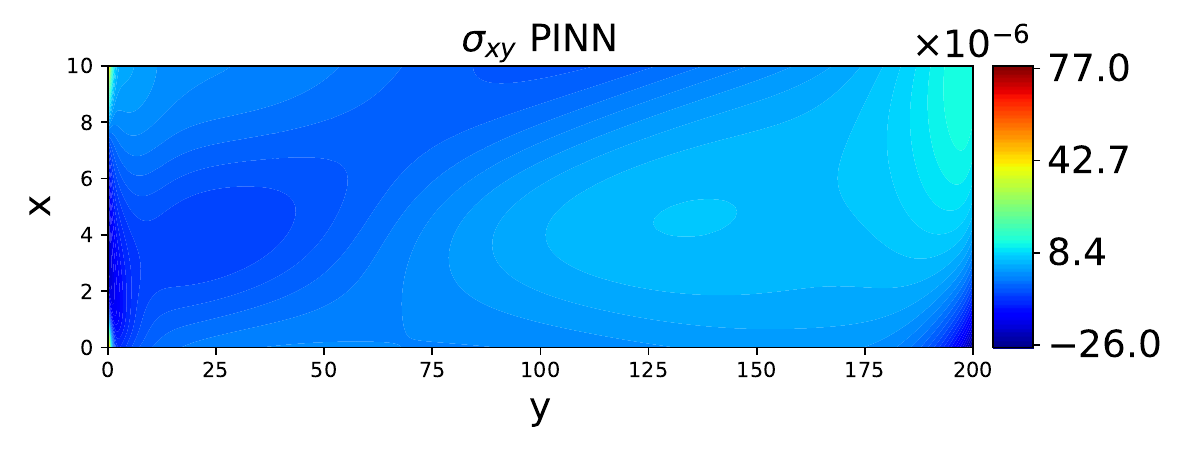}
\caption{}
\end{subfigure}
\begin{subfigure}[t]{0.33\textwidth}
\includegraphics[width=\textwidth,height=\textheight,keepaspectratio]{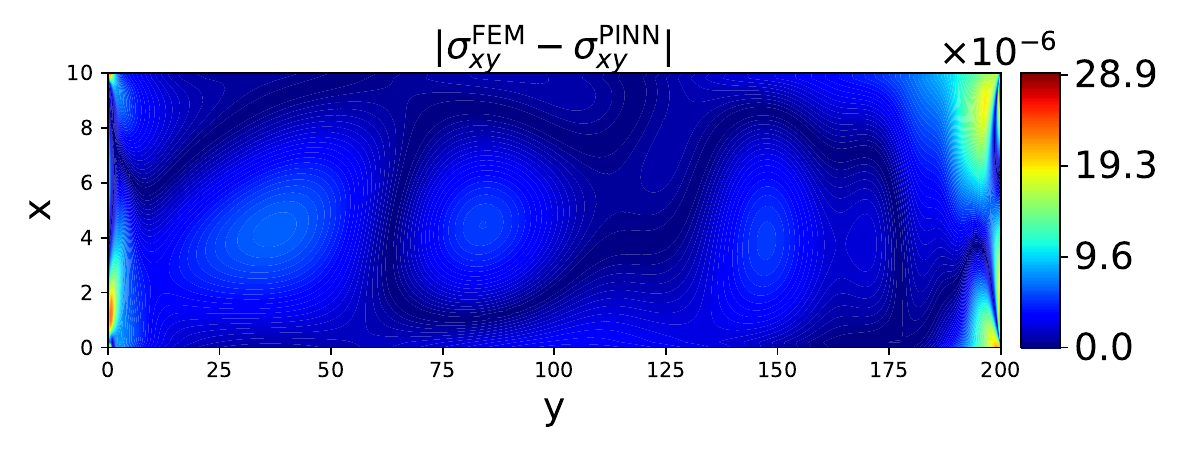}
\caption{}
\end{subfigure}
\caption{Contour plots of the quantities of interest at $t=1.72$ s for the material property $\mu=0.075$ MPa. Plots in (a) denote FEM solutions. Plots in (b) denote PINN solutions. Plots in (c) denotes the absolute error.}
\label{fig:contours_surr}
\end{figure}

\begin{figure*}
\begin{subfigure}[t]{0.49\textwidth}
\includegraphics[width=\textwidth,height=\textheight,keepaspectratio]{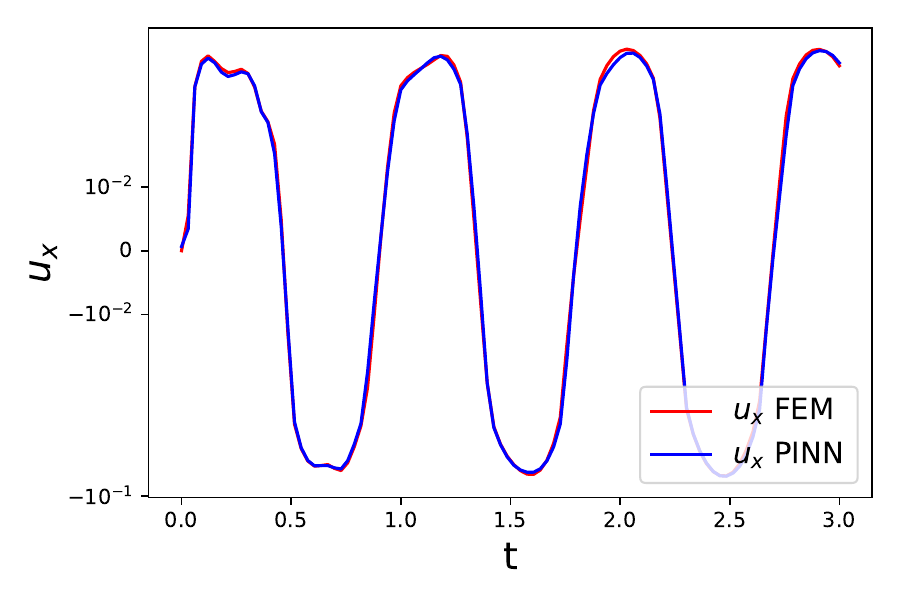}
\caption{}
\end{subfigure}
\begin{subfigure}[t]{0.49\textwidth}
\includegraphics[width=\textwidth,height=\textheight,keepaspectratio]{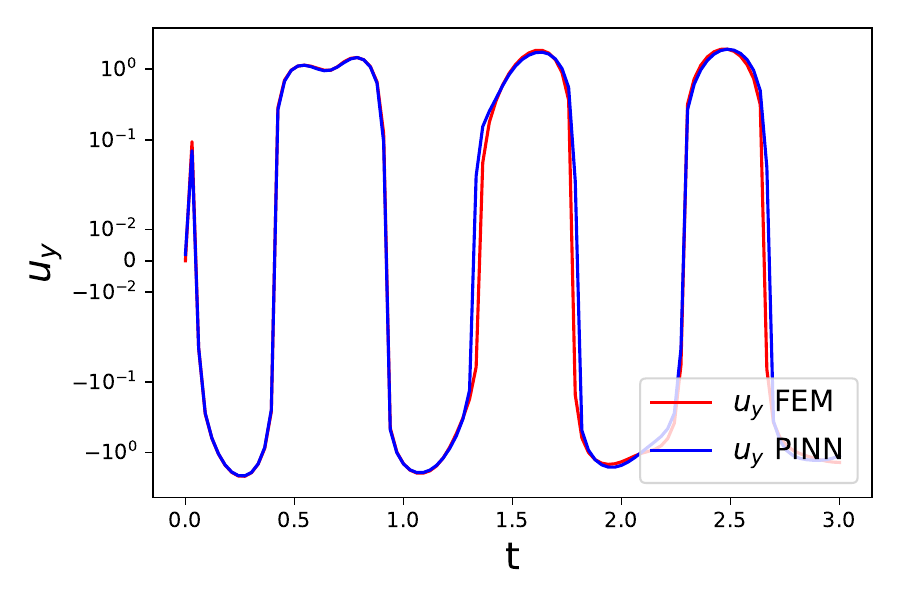}
\caption{}
\end{subfigure}
\begin{subfigure}[t]{0.33\textwidth}
\includegraphics[width=\textwidth,height=\textheight,keepaspectratio]{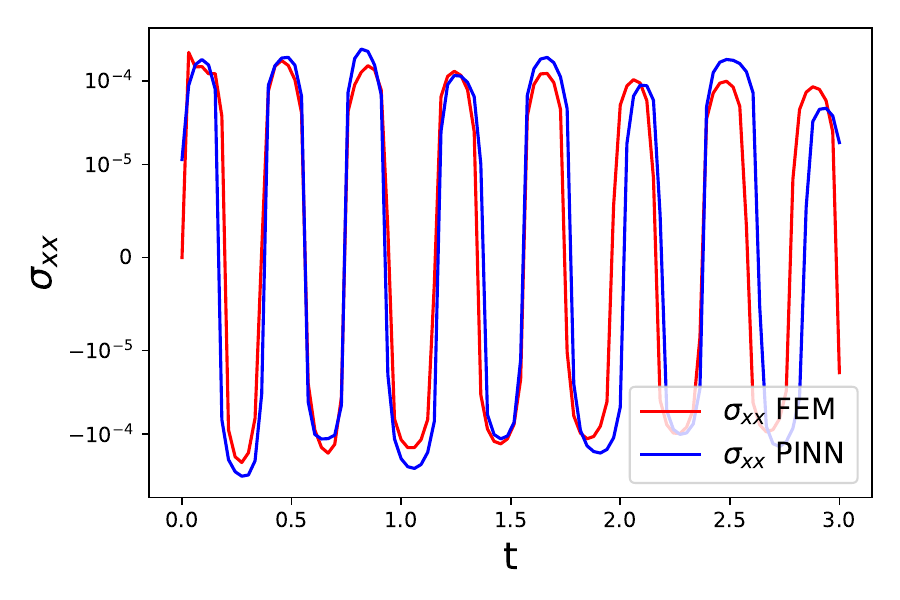}
\caption{}
\end{subfigure}
\begin{subfigure}[t]{0.33\textwidth}
\includegraphics[width=\textwidth,height=\textheight,keepaspectratio]{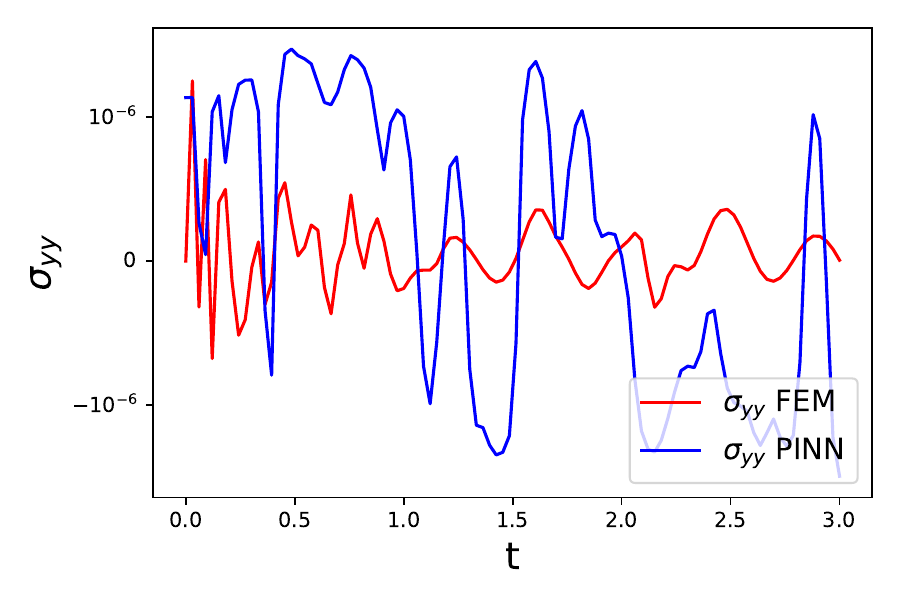}
\caption{}
\end{subfigure}
\begin{subfigure}[t]{0.33\textwidth}
\includegraphics[width=\textwidth,height=\textheight,keepaspectratio]{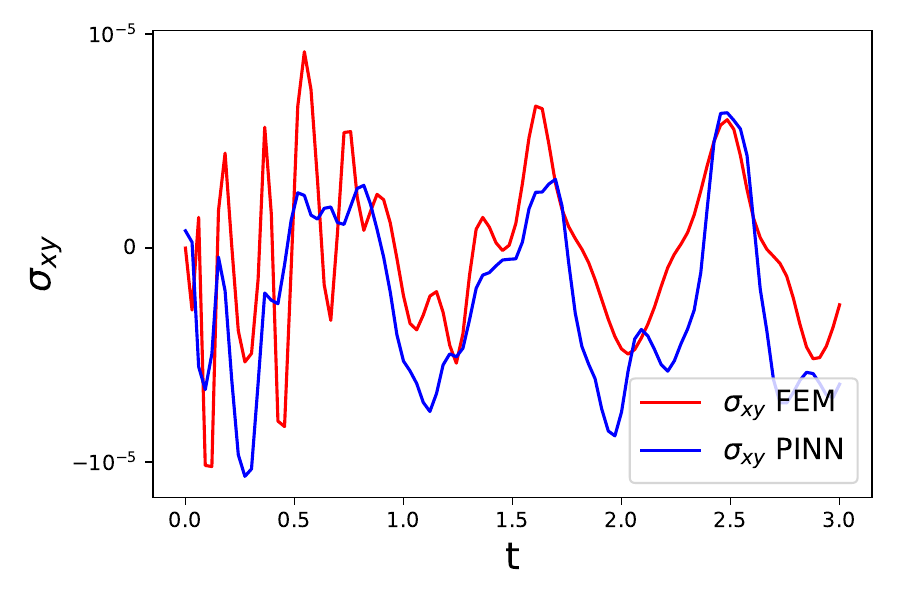}
\caption{}
\end{subfigure}
\caption{Line plots of the quantities of interest at $(x=120 \ \text{mm},\ y=8 \ \text{mm})$ for material property $\mu=0.075$ MPa. Plots in (a) to (e) show the comparison of displacements and stresses obtained by PINN with respect to the FEM solutions.}
\label{fig:time_series_surr}
\end{figure*}

Due to practical constraints on data procurement via experimental measurements, traditionally, numerical simulations are a mainstay for problems in solid mechanics. However, in this context of material property estimation, one resorts to performing repeated simulations for various iterative combinations of the parameters. This iterative search ends when one finds the parameters for which the experimental data matches with the simulation. Thus, one incurs high amount of computational cost. In this section, we will show how a PINN model can be used to circumvent this issue. We consider the same problem of a beam with an oscillating excitation at the free end as shown in Fig. \ref{fig:schema1}. However, in this case, we train the PINN model with simulation data of six different values of the material property $\mu$ while keeping $\lambda=0.533334$ MPa constant. The parameter $\mu$ is also given as additional input feature to the network along with spatial and temporal coordinates as $(x,y,t,\mu)$.  The training values chosen for $\mu$ are shown in the six five rows of Table \ref{tab:surrogate_errors}. In the same table, the bottom three rows below a horizontal line denote the testing space $\mu$ values.For training, in space, we choose $23.9$ \% of the boundaries data and in time we choose all time instances of FEM simulation. This training data  accounts for nearly $0.98$ \% of entire simulation data. We try to constrain model toward learning solution based on equations properly, in order to fulfill that we give more weightage to the equation loss by selecting $\alpha_1 = \alpha_2 = 10^3$ so as to make equation loss tends towards even lower order of magnitudes.  Also, we show in Table \ref{tab:surrogate_errors} the error values $\epsilon$ for $u_x$ and $u_y$. We observe that the testing errors are one order higher than the training errors. However, they are still low enough to obtain physically meaningful solutions. In Fig. \ref{fig:contours_surr}, we show the contour plots from the PINN model and FEM for testing $\mu=0.075$ MPa. We notice a high accuracy in the prediction of $u_x$ and $u_y$ from PINN. Even if stresses are not exactly matching to the FEM solution, they are of the same order of magnitude. Furthermore, in Fig. \ref{fig:time_series_surr}, we show the error $\epsilon$ as a function of time for a particular location in space. We observe again a very close match between PINN and FEM for the displacements. Barring the training time for PINN, the prediction based on PINN for new parameters $\mu$ is extremely fast, usually seconds. For the same problem where $\mu=0.075$ MPa, the FEM simulation takes approximately $20$ minutes on a standalone Intel Core i7 CPU. Hence, the PINN model easily outperforms the FEM simulation in terms of computational cost while maintaining reasonable accuracy.

\begin{table*}
\caption{\label{tab:table1} Error values for different training and testing values of the material parameter $\mu$ with respect to FEM solution. Bottom three rows denote testing space of the parameters.}
\begin{ruledtabular}
\begin{tabular}{lccc}
Parameter $\mu \ \text{(MPa)}$  & $\epsilon_{u_x}$ & $\epsilon_{u_y}$ \\
\hline
$0.05$ &  $4.62\times 10^{-3}$ & $3.2 \times 10^{-3}$  \\
$0.07$ &  $4.22 \times 10^{-3}$ & $2.17 \times 10^{-3}$ \\
$0.08$ &  $3.47 \times 10^{-3}$ & $1.76 \times 10^{-3}$  \\
$0.1$ &  $3.83 \times 10^{-3}$ & $1.85 \times 10^{-3}$  \\
$0.2$ &  $3.65 \times 10^{-3}$ & $1.31 \times 10^{-3}$  \\
$0.4$ &  $4.85 \times 10^{-3}$ & $1.35 \times 10^{-3}$  \\
\hline
$0.06$ &  $2.67 \times 10^{-2}$ & $2.07 \times 10^{-2}$ \\
$0.075$ &  $1.25 \times 10^{-2}$ & $1.07 \times 10^{-2}$ \\
$0.3$ &  $3.31 \times 10^{-2}$ & $1.93 \times 10^{-2}$
\end{tabular}
\label{tab:surrogate_errors}
\end{ruledtabular}
\end{table*}

\subsection{\label{3D-cantilever} Extension to a 3D surrogate model}
\label{Sec:3D}

\subsubsection{Problem setup (3D)}
\begin{figure}
\centering
\includegraphics[width=0.8\textwidth,height=0.5\textheight,keepaspectratio]{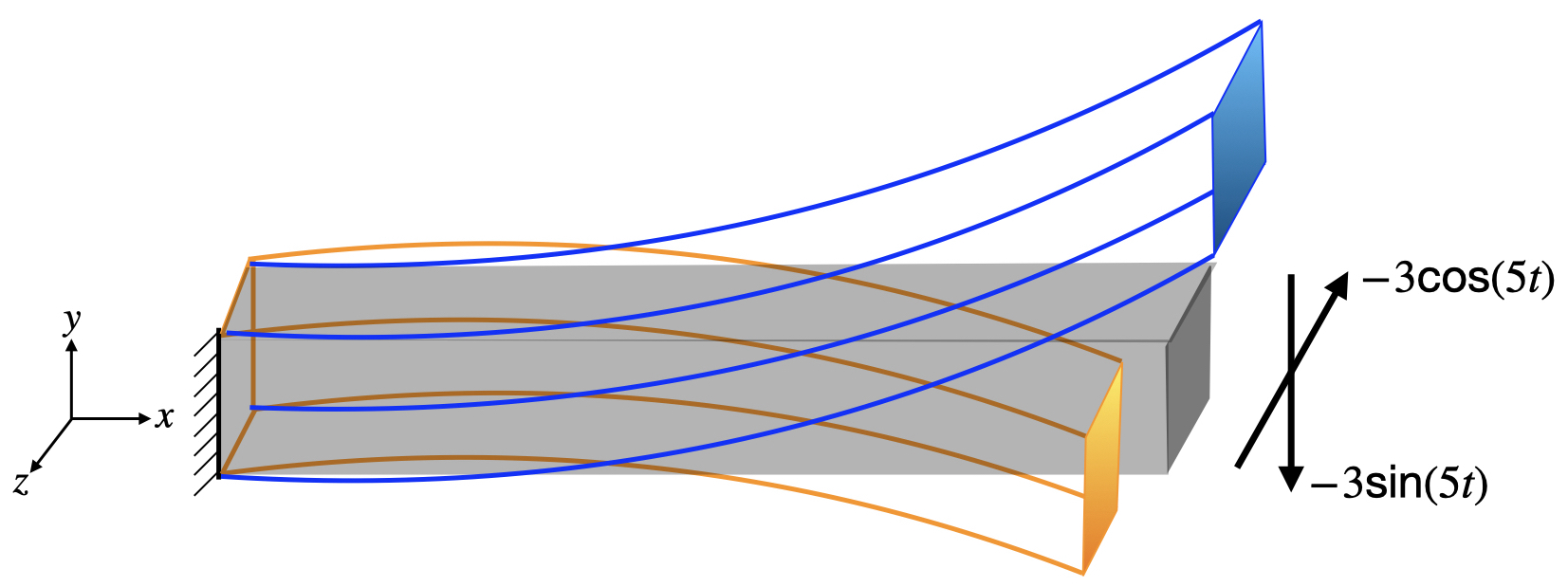}
\caption{\label{fig:schema2} \markup{Schematic of the 3D cantilever beam under an oscillating excitation at the free end. For clarity, blue and orange beams show how the beam moves in 3D at two different instances in time.}}
\end{figure}

A schematic of the problem is shown in Fig. \ref{fig:schema2} where the 3D cantilever beam is under oscillating excitation at the free end in both $y$ and $z$ axes. We consider beam of size 200 mm $\times$ 10 mm $\times$ 10 mm  with 
$ x \in [0,200]  \ \text{mm}$, $y \in [0,10] \ \text{mm}$ , $z \in [0,10] \ \text{mm}$. We assume a linearly elastic material with a density of $\rho=0.92 \times 10^{-6} \ \text{kg}/\text{mm}^3$. The material properties are considered to be the same as considered for the 2D surrogate model. For the FEM simulation, spatially, the solution is computed on a grid with $24321$ nodes of which $8202$ nodes belong to the boundary surfaces. The total simulation time is set for $3$ seconds. 

\subsubsection{Results and discussion (3D)}
\begin{figure}
\begin{subfigure}[t]{0.33\textwidth}
\includegraphics[width=\textwidth,height=\textheight,keepaspectratio]{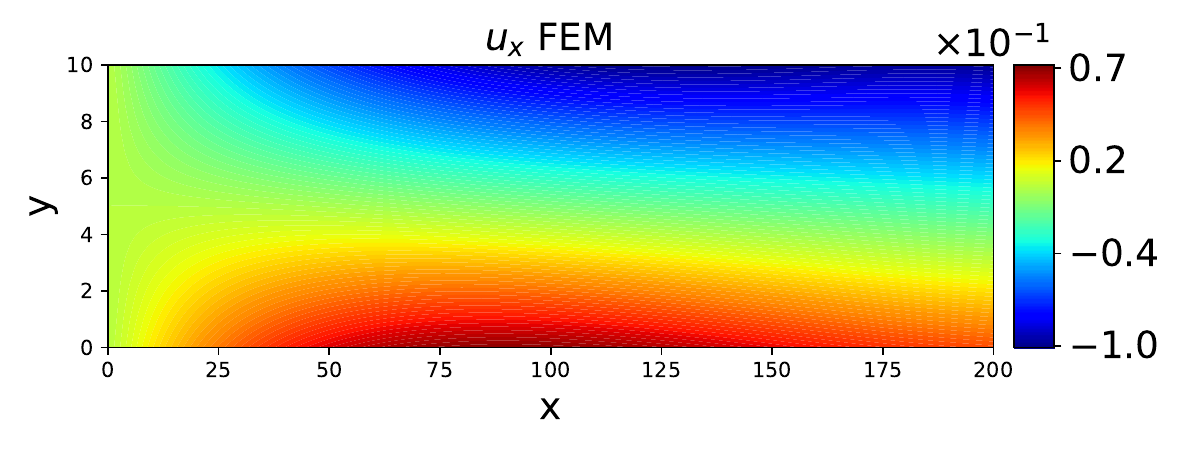}
\end{subfigure}
\begin{subfigure}[t]{0.33\textwidth}
\includegraphics[width=\textwidth,height=\textheight,keepaspectratio]{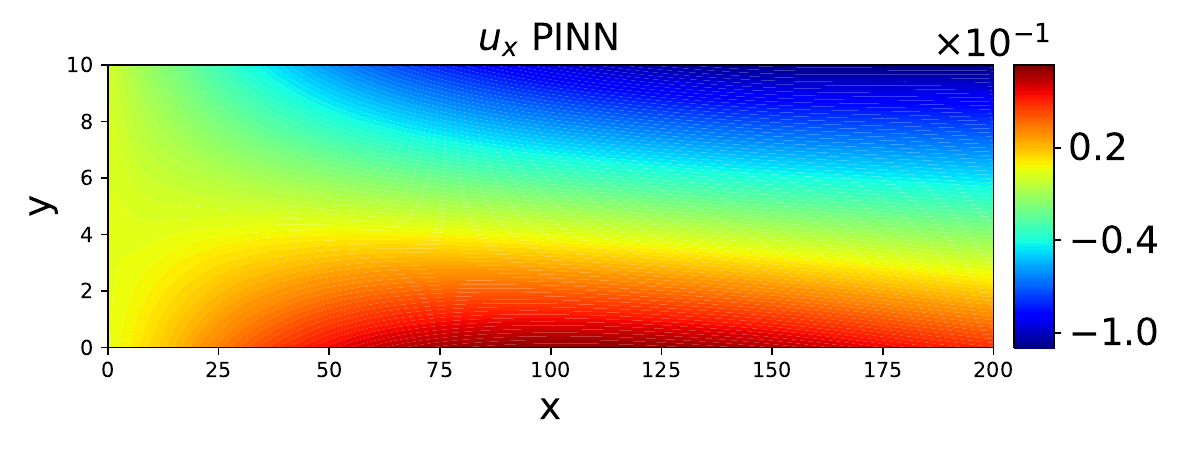}
\end{subfigure}
\begin{subfigure}[t]{0.33\textwidth}
\includegraphics[width=\textwidth,height=\textheight,keepaspectratio]{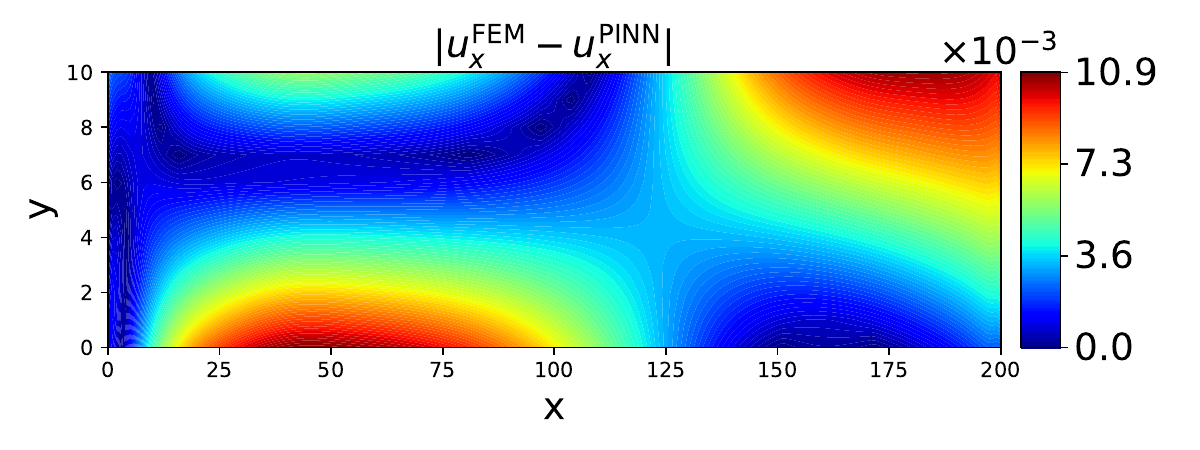}
\end{subfigure}
\begin{subfigure}[t]{0.33\textwidth}
\includegraphics[width=\textwidth,height=\textheight,keepaspectratio]{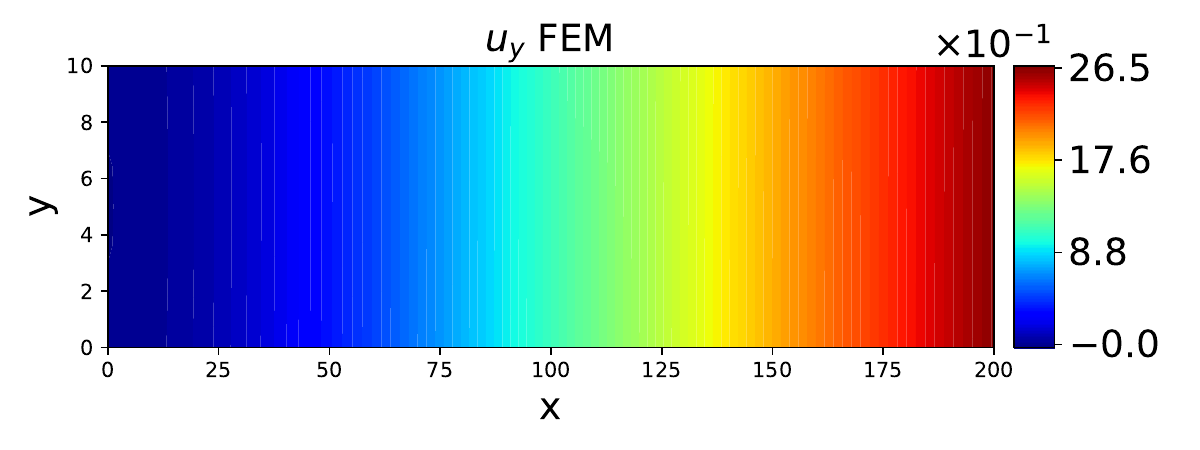}
\end{subfigure}
\begin{subfigure}[t]{0.33\textwidth}
\includegraphics[width=\textwidth,height=\textheight,keepaspectratio]{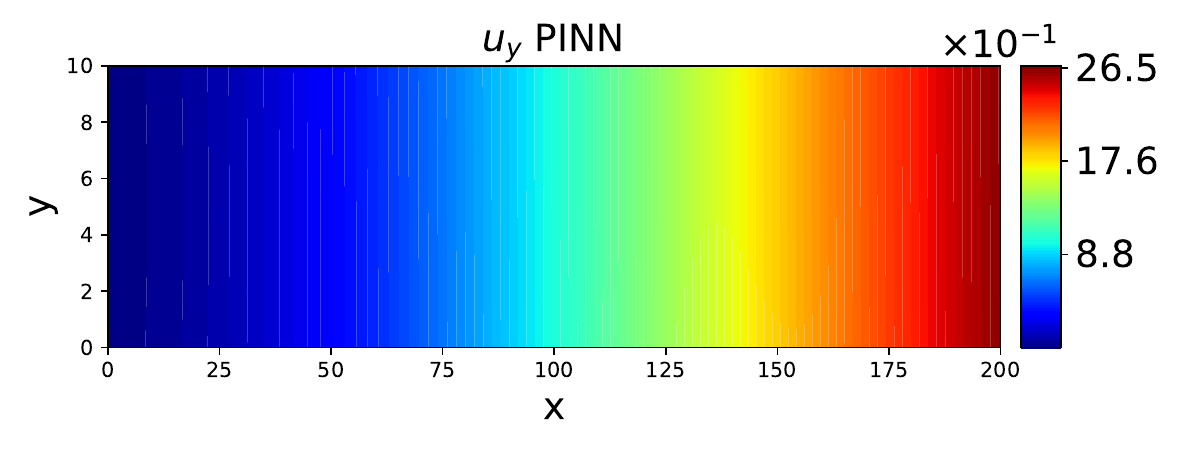}
\end{subfigure}
\begin{subfigure}[t]{0.33\textwidth}
\includegraphics[width=\textwidth,height=\textheight,keepaspectratio]{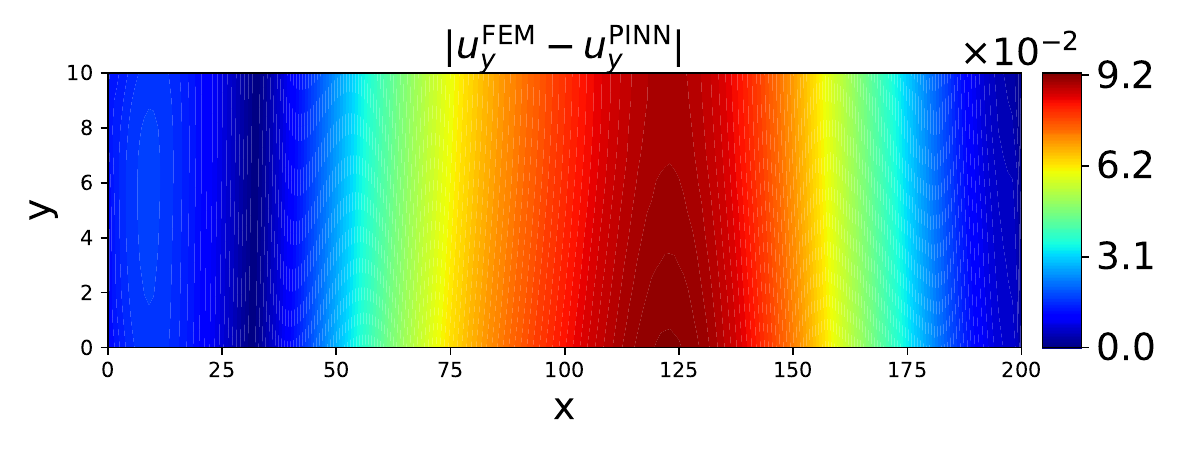}
\end{subfigure}
\begin{subfigure}[t]{0.33\textwidth}
\includegraphics[width=\textwidth,height=\textheight,keepaspectratio]{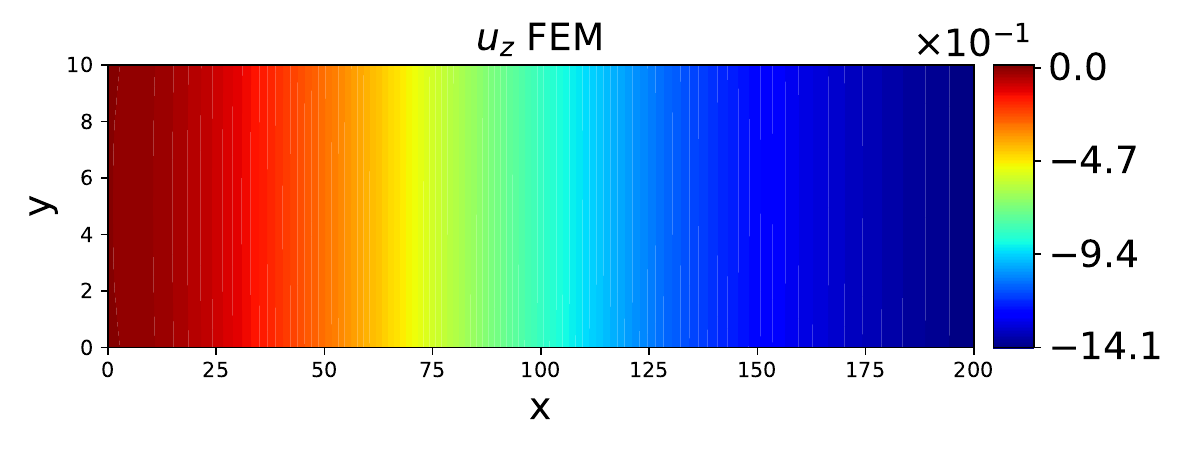}
\end{subfigure}
\begin{subfigure}[t]{0.33\textwidth}
\includegraphics[width=\textwidth,height=\textheight,keepaspectratio]{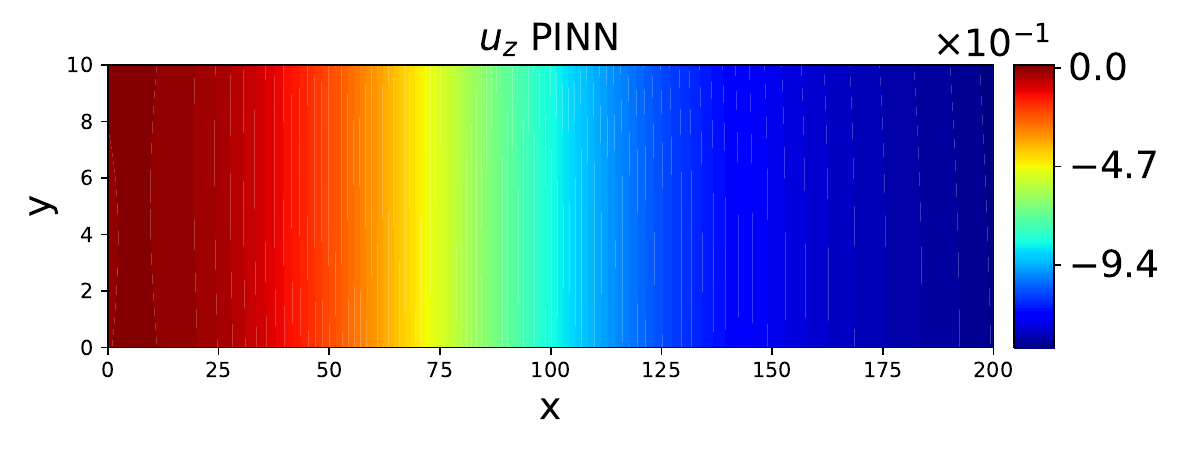}
\end{subfigure}
\begin{subfigure}[t]{0.33\textwidth}
\includegraphics[width=\textwidth,height=\textheight,keepaspectratio]{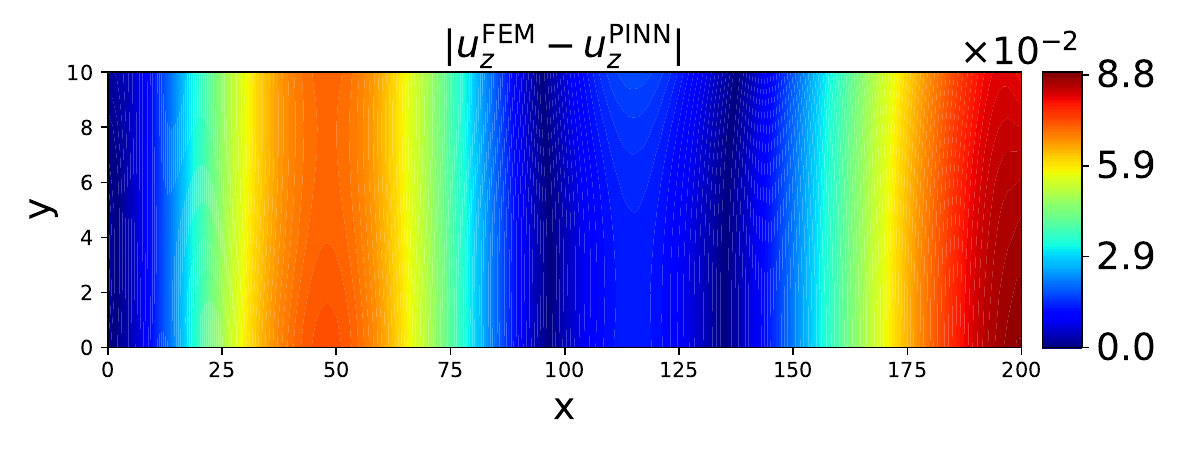}
\end{subfigure}
\caption{Contour plots of the quantities of interest in the $xy$-plane at $z=5 \ mm$ and $t=2.29$ s for the material property $\mu=0.075$ MPa. Plots in (a) denote FEM solutions. Plots in (b) denote PINN solutions. Plots in (c) denotes the absolute error.}
\label{fig:contours_surr_3d}
\end{figure}

\begin{figure}
\begin{subfigure}[t]{0.33\textwidth}
\includegraphics[width=\textwidth,height=\textheight,keepaspectratio]{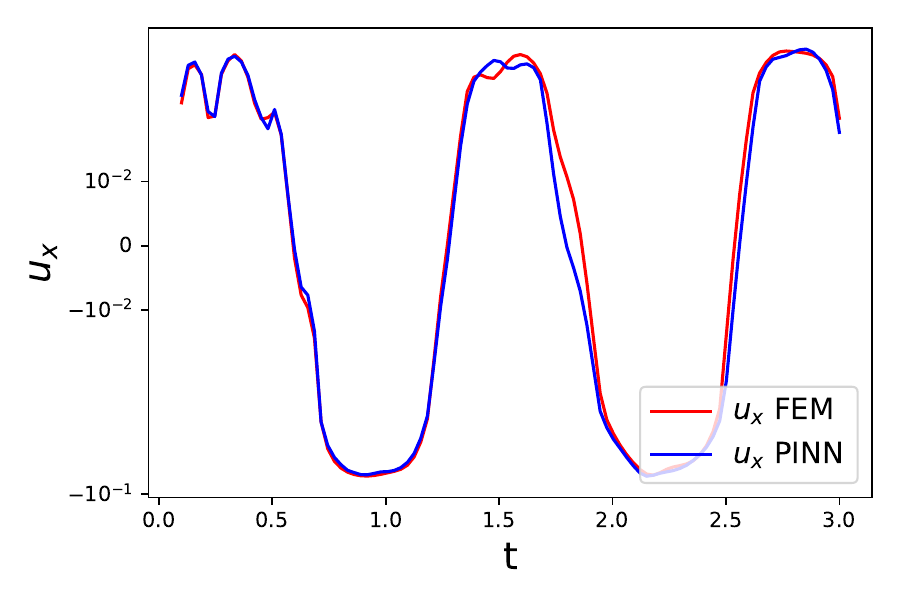}
\caption{}
\end{subfigure}
\begin{subfigure}[t]{0.33\textwidth}
\includegraphics[width=\textwidth,height=\textheight,keepaspectratio]{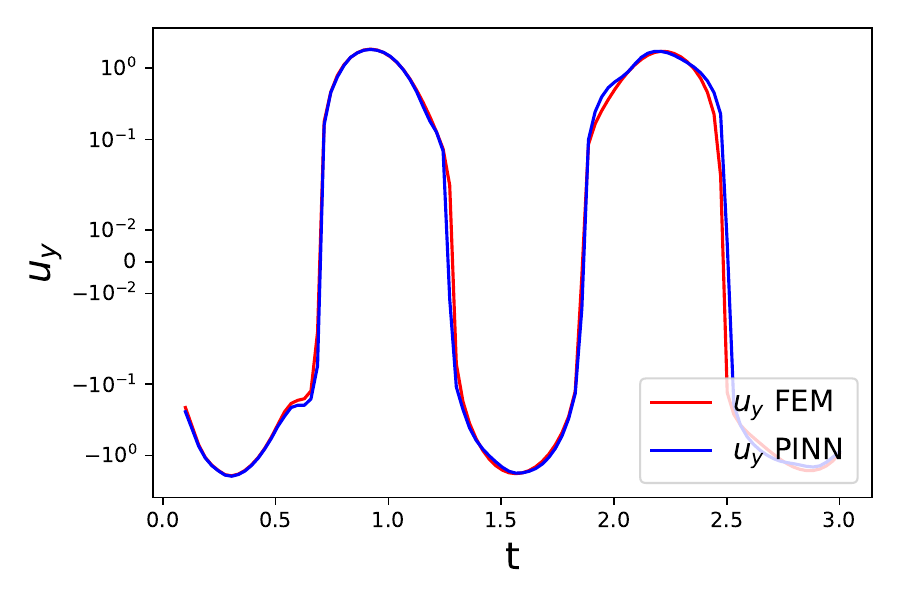}
\caption{}
\end{subfigure}
\begin{subfigure}[t]{0.33\textwidth}
\includegraphics[width=\textwidth,height=\textheight,keepaspectratio]{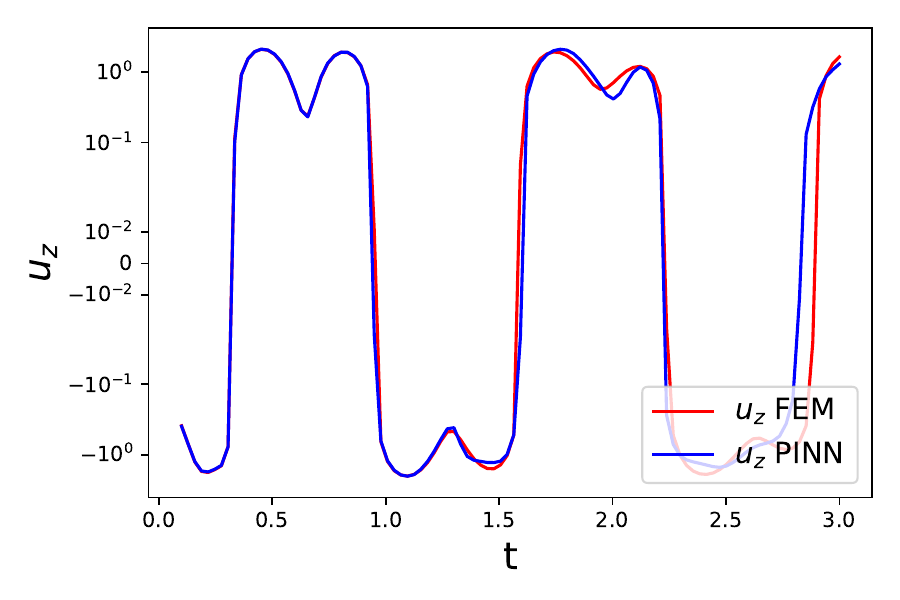}
\caption{}
\end{subfigure}
\begin{subfigure}[t]{0.33\textwidth}
\includegraphics[width=\textwidth,height=\textheight,keepaspectratio]{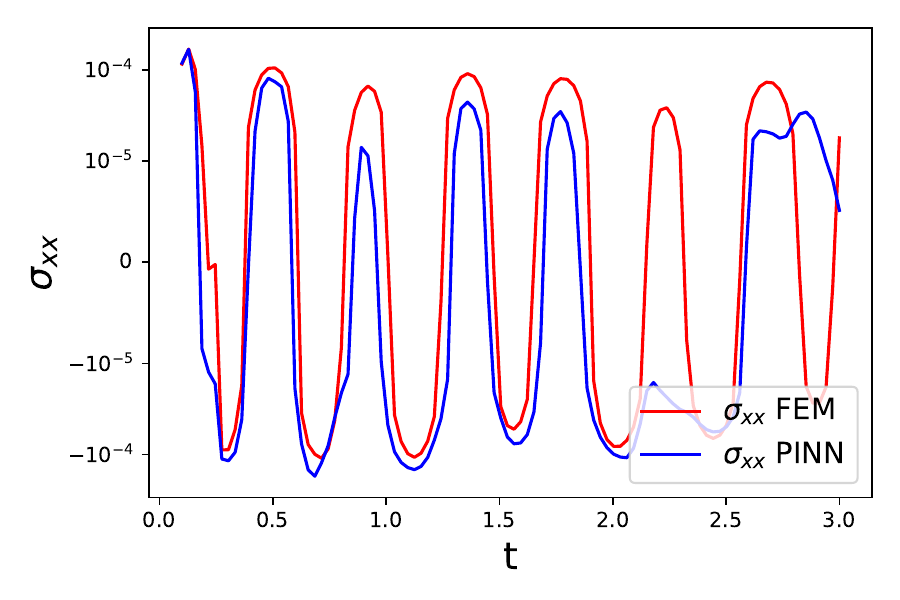}
\caption{}
\end{subfigure}
\begin{subfigure}[t]{0.33\textwidth}
\includegraphics[width=\textwidth,height=\textheight,keepaspectratio]{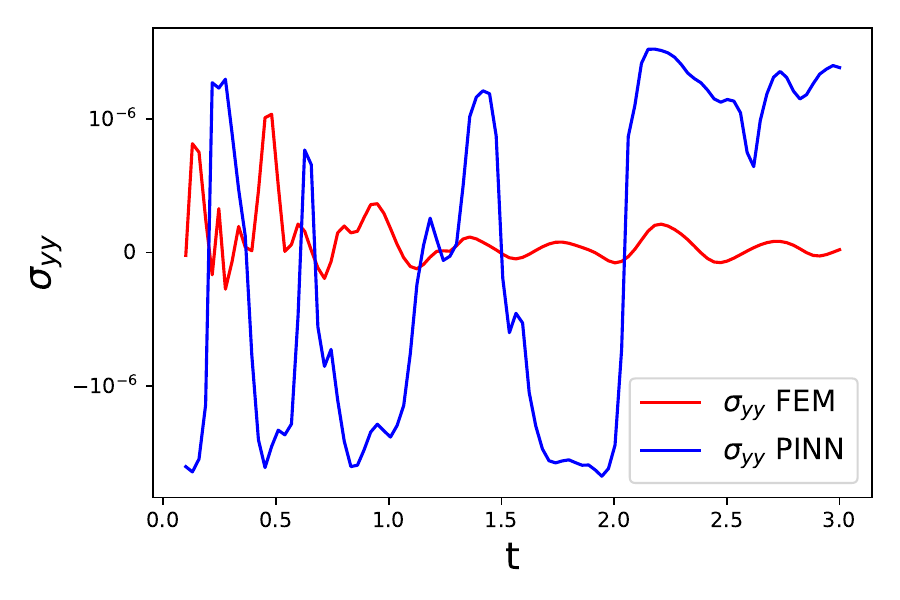}
\caption{}
\end{subfigure}
\begin{subfigure}[t]{0.33\textwidth}
\includegraphics[width=\textwidth,height=\textheight,keepaspectratio]{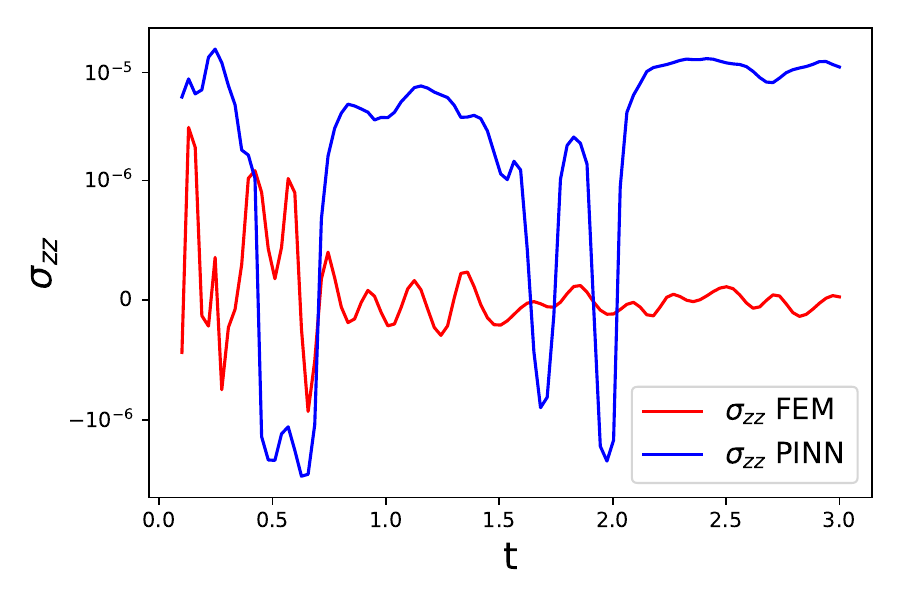}
\caption{}
\end{subfigure}
\begin{subfigure}[t]{0.33\textwidth}
\includegraphics[width=\textwidth,height=\textheight,keepaspectratio]{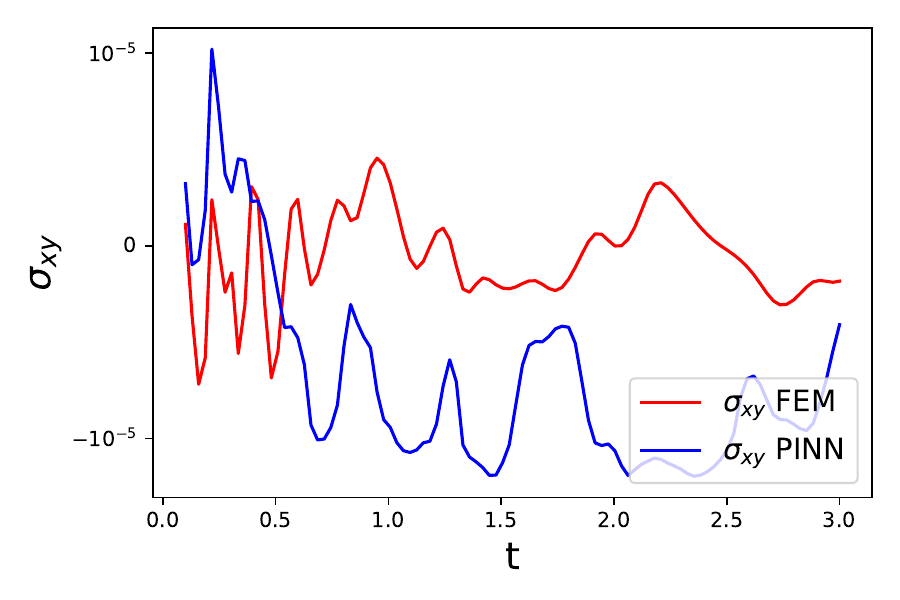}
\caption{}
\end{subfigure}
\begin{subfigure}[t]{0.33\textwidth}
\includegraphics[width=\textwidth,height=\textheight,keepaspectratio]{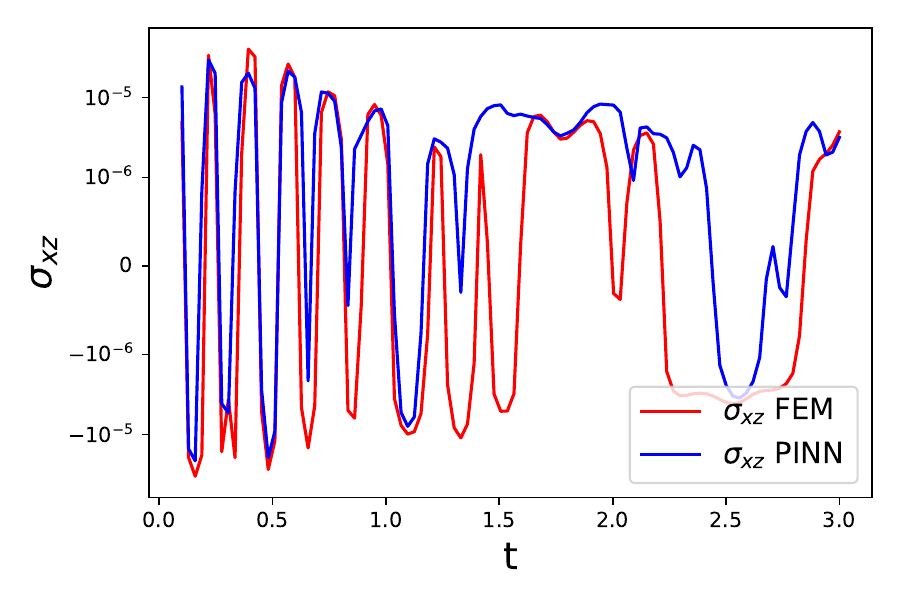}
\caption{}
\end{subfigure}
\begin{subfigure}[t]{0.33\textwidth}
\includegraphics[width=\textwidth,height=\textheight,keepaspectratio]{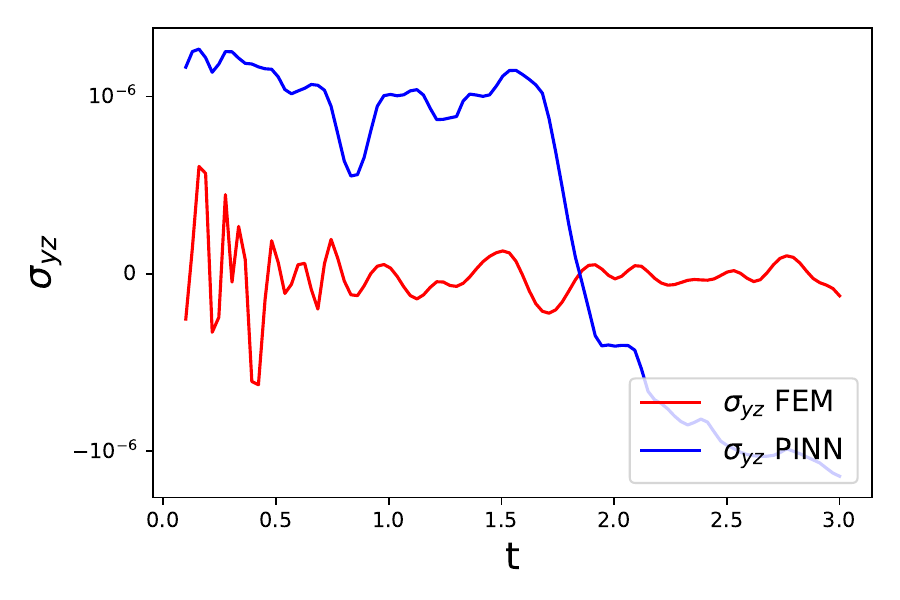}
\caption{}
\end{subfigure}
\caption{Line plots of the quantities of interest at $(x=120 \ \text{mm}, \ y=8 \ \text{mm}, \ z=5 \ \text{mm})$ for material property $\mu=0.075$ MPa. Plots in (a) to (e) show the comparison of displacements and stresses obtained by PINN with respect to the FEM solutions.}
\label{fig:time_series_surr_3d}
\end{figure}

 We apply the PINN Framework as surrogate model for 3D. We scale the quantities as mentioned in section3. As for the neural network architecture, we use a single-network where it takes the scaled quantities as  inputs $(x^*,y^*,z^*,t^*, \mu^*)$ and outputs scaled quantities $(u_x^*, u_y^*, u_z^*, \sigma_{xx}^*,\sigma_{yy}^*, \sigma_{zz}^*,\sigma_{xy}^*,\sigma_{yz}^*,\sigma_{xz}^*)$.  
 Once the model is trained we rescale the neural network outputs to get actual physical quantities.  Also, the values for training and testing material parameter $\mu$ is given in Table \ref{tab:surrogate_errors_3D}. For training, in space, we take 5.48 \% of the boundary surfaces data and in time we choose all time instances of FEM simulation. This training data accounts for nearly 1.85 \% of entire simulation data. We consider the network architecture consisting of 256 neurons per hidden layer and 4 hidden layers.  We train the model for a total of $1500$ {\markup{epochs}}. For the first $500$ {\markup{epochs}}, we use a learning rate of $\eta=0.001$. For the next $1000$ {\markup{epochs}}, we keep $\eta=0.0001$. The batch size is considered to be 225 and the number of collocation points to be 1000 at each gradient descent step. The loss functions in this case are defined as,
\begin{align}
\mathcal{L}_{tot} = \Lossdata +  \Losseqn,
\end{align}

\begin{multline}
\Lossdata =  \|{u_x^*} - {u_x^*}^{d}\|_{data} + \|{u_y^*} - {u_y^*}^{d}||_{data} + \|{u_z^*} - {u_z^*}^{d}||_{data} + 
\|{\sigma_{xx}^*} - {\sigma_{xx}^*}^{d}\|_{data} + \|{\sigma_{yy}^*} - {\sigma_{yy}^*}^{d}\|_{data} + \\ \|{\sigma_{zz}^*} - {\sigma_{zz}^*}^{d}\|_{data} + \|{\sigma_{xy}^*} - {\sigma_{xy}^*}^{d}\|_{data} 
+  \|{\sigma_{yz}^*} - {\sigma_{yz}^*}^{d}\|_{data} + 
\|{\sigma_{xz}^*} - {\sigma_{xz}^*}^{d}\|_{data}
\label{eqn:loss_data_expand} 
\end{multline}

\begin{equation}
\begin{split}
{  \mathcal{L}_{eqn} = \alpha_1 \mathcal{L}_{x} + \alpha_2 \mathcal{L}_{y} + + \alpha_3 \mathcal{L}_{z} + \alpha_4 \mathcal{L}_{xx} + \alpha_5 \mathcal{L}_{yy} + \alpha_6 \mathcal{L}_{zz} + 
\alpha_7 \mathcal{L}_{xy} + \alpha_8 \mathcal{L}_{yz} + 
\alpha_9 \mathcal{L}_{xz}
}
\end{split}
\label{loss_eqn_alpha}
\end{equation}

\begin{equation}
\begin{split}
{
\mathcal{L}_{x} =  \biggr\| \frac{\partial \sigma_{xx}^*}{\partial x^*} +   \frac{{\sigma_{xy}}_c}{{\sigma_{xx}}_c}  \frac{\partial \sigma_{xy}^*}{\partial y^*} + 
\frac{{\sigma_{xz}}_c}{{\sigma_{xx}}_c}  \frac{\partial \sigma_{xz}^*}{\partial z^*}
- \frac{ {u_x}_c \rho_c  l_c}{ {\sigma_{xx}}_c t_c^2 }  \rho^* \frac{\partial^2 u_x^*}{\partial {t^*}^2}  \biggr\|_{eqn}
 }
\end{split}
\end{equation}

\begin{equation}
\begin{split}
{
\mathcal{L}_{y} =  \biggr\|  \frac{{\sigma_{xy}}_c}{{\sigma_{xx}}_c} \frac{\partial \sigma_{xy}^*}{\partial x^*} +  \frac{{\sigma_{yy}}_c}{{\sigma_{xx}}_c}  \frac{\partial \sigma_{yy}^*}{\partial y^*} + 
\frac{{\sigma_{yz}}_c}{{\sigma_{xx}}_c} \frac{\partial \sigma_{yz}^*}{\partial z^*} - \frac{ {u_y}_c \rho_c  l_c}{ {\sigma_{xx}}_c t_c^2 }  \rho^* \frac{\partial^2 u_y^*}{\partial {t^*}^2}  \biggr\|_{eqn}
 }
\end{split}
\end{equation}

\begin{equation}
\begin{split}
{
\mathcal{L}_{z} =  \biggr\|  \frac{{\sigma_{xz}}_c}{{\sigma_{xx}}_c} \frac{\partial \sigma_{xz}^*}{\partial x^*} +  \frac{{\sigma_{yz}}_c}{{\sigma_{xx}}_c}  \frac{\partial \sigma_{yz}^*}{\partial y^*} + 
\frac{{\sigma_{zz}}_c}{{\sigma_{xx}}_c} \frac{\partial \sigma_{zz}^*}{\partial z^*} - \frac{ {u_z}_c \rho_c  l_c}{ {\sigma_{xx}}_c t_c^2 }  \rho^* \frac{\partial^2 u_z^*}{\partial {t^*}^2}  \biggr\|_{eqn}
 }
\end{split}
\end{equation}

\begin{equation}
\begin{split}
{
\mathcal{L}_{xx} =\biggr\|  \frac{ {\sigma_{xx}}_c l_c }{ {u_z}_c \lambda_c}  \sigma_{xx}^* - \frac{ {u_x}_c }{ {u_z}_c } (\lambda^* + 2 \mu^*) \frac{\partial u_x^* }{\partial x^*} -   \frac{ {u_y}_c }{ {u_z}_c } \lambda^* \frac{\partial {u_y}^*}{\partial y^*} - \lambda^* \frac{\partial {u_z}^*}{\partial z^*} \biggr\|_{eqn}
}  
\end{split}
\end{equation}

\begin{equation}
\begin{split}
{
\mathcal{L}_{yy} =\biggr\|  \frac{ {\sigma_{yy}}_c l_c }{ {u_z}_c \lambda_c}  \sigma_{yy}^* - \frac{ {u_x}_c }{ {u_z}_c } \lambda^* \frac{\partial u_x^* }{\partial x^*} -   \frac{ {u_y}_c }{ {u_z}_c } (\lambda^* + 2 \mu^*) \frac{\partial {u_y}^*}{\partial y^*} - \lambda^* \frac{\partial {u_z}^*}{\partial z^*} \biggr\|_{eqn}
}  
\end{split}
\end{equation}

\begin{equation}
\begin{split}
{
\mathcal{L}_{zz} =\biggr\|  \frac{ {\sigma_{zz}}_c l_c }{ {u_z}_c \lambda_c}  \sigma_{yy}^* - \frac{ {u_x}_c }{ {u_z}_c } \lambda^* \frac{\partial u_x^* }{\partial x^*} -   \frac{ {u_y}_c }{ {u_z}_c } \lambda^* \frac{\partial {u_y}^*}{\partial y^*} - (\lambda^* + 2 \mu^*) \frac{\partial {u_z}^*}{\partial z^*} \biggr\|_{eqn}
}  
\end{split}
\end{equation}

\begin{equation}
\begin{split}
{
\mathcal{L}_{xy} =\biggr\|  \frac{ {\sigma_{xy}}_c l_c }{ {u_z}_c \lambda_c}  \sigma_{xy}^* -  \mu^*\biggl( \frac{ {u_y}_c }{{u_z}_c} \frac{\partial u_y^*}{\partial x^*} + \frac{ {u_x}_c }{{u_z}_c} \frac{\partial u_x^*}{\partial y^*} \biggr)  \biggr\|_{eqn}  
}  
\end{split}
\end{equation}

\begin{equation}
\begin{split}
{
\mathcal{L}_{yz} =\biggr\|  \frac{ {\sigma_{yz}}_c l_c }{ {u_z}_c \lambda_c}  \sigma_{yz}^* -  \mu^*\biggl( \frac{ {u_y}_c }{{u_z}_c} \frac{\partial u_y^*}{\partial z^*} + \frac{\partial u_z^*}{\partial y^*} \biggr)  \biggr\|_{eqn}  
}  
\end{split}
\end{equation}

\begin{equation}
\begin{split}
{
\mathcal{L}_{xz} =\biggr\|  \frac{ {\sigma_{xz}}_c l_c }{ {u_z}_c \lambda_c}  \sigma_{xz}^* -  \mu^*\biggl( \frac{ {u_x}_c }{{u_z}_c} \frac{\partial u_x^*}{\partial z^*} + \frac{\partial u_z^*}{\partial x^*} \biggr)  \biggr\|_{eqn}  
}  
\end{split}
\end{equation}
We consider the equation loss coefficients $\alpha_1$ to $\alpha_9$ as unity during the training. We discuss the results from the PINN model for an unseen or testing material parameters $\mu$. We show in Fig. \ref{fig:contours_surr_3d} the comparison of displacements with respect to the FEM results for $\mu = 0.075$ MPa. We observe a close visual and quantitative match between the PINN and FEM solutions. For $u_x$, the absolute error is maximum at top right and bottom left sides of the beam. For $u_y$, the absolute error is maximum close to the mid section of the beam along $x$ direction. For $u_z$ it is maximum at the right end of the beam. The reason we choose to show only displacements instead of the stresses is because of a practical reason that in real industrial experiments, displacements can be measured accurately using sensors on the material but stress measurements are bit complicated and sometimes not possible. We compute NRMSE error $\epsilon$ at 51 $\times$ 11 $\times$ 11 $\times$ 100 spatio-temporal points which corresponds to 51 $\times$ 11 $\times$ 11 sparse spatial grid points and 100 time instants. In Table \ref{tab:surrogate_errors_3D}, we show the numerical value of the errors in the displacements for both training and testing values of the parameter $\mu$. The bottom three rows comprise of the testing values. We observe the testing errors in displacements are one order higher as compared to the training ones. This is still acceptable because the testing errors are of the order $\mathcal{O}(10^{-2})$ which is quite low. For an inclusive view of the results, we also show in Fig. \ref{fig:time_series_surr_3d} the time series of the spatial errors in displacements and stresses calculated at one random location in the 3D beam. We observe the displacements to be matching closely. As for the stresses, they are not following the trend of the FEM solution. This is because the neural network has a bias towards learning lower frequency functions better. However, the predicted stress are of the same order as that of the FEM solution. For an unseen or testing value of the parameter $\mu$, worthy to note is that the computational cost for such a dynamic 3D simulation is mere seconds using PINN whereas it takes approximately $20$ minutes using FEM on a standalone Core i7 CPU. Using such a deep learning framework, if one has to perform $N$ number of simulations for $N$ parameters, that too on a complicated geometry, one can rely on a trained PINN model to provide extremely fast solutions with reasonable accuracy.

\begin{table*}
\caption{\label{tab:table1} Error values for different training and testing values of the material parameter $\mu$ with respect to FEM solution. Bottom three rows denote testing space of the parameters. (3D problem) }
\begin{ruledtabular}
\begin{tabular}{lccc}
Parameter $\mu$ (MPa) & $\epsilon_{u_x}$ & $\epsilon_{u_y}$ & $\epsilon_{u_z}$ \\
\hline
$0.05$ &  $3.1\times 10^{-3}$ & $3.6 \times 10^{-3}$  & $3.6 \times 10^{-3}$ \\
$0.07$ &  $4.4 \times 10^{-3}$ & $3.3 \times 10^{-3}$ & $3.5 \times 10^{-3}$ \\
$0.08$ &  $4.2 \times 10^{-3}$ & $3.1 \times 10^{-3}$ & $3.7 \times 10^{-3}$ \\
$0.1$ &  $3.8 \times 10^{-3}$ & $2.1 \times 10^{-3}$ & $1.8 \times 10^{-3}$ \\
$0.2$ &  $3.5 \times 10^{-3}$ & $1 \times 10^{-3}$ & $1.1 \times 10^{-3}$  \\
$0.4$ &  $3.3 \times 10^{-3}$ & $6 \times 10^{-4}$ & $7 \times 10^{-4}$ \\
\hline
$0.06$ &  $4.6 \times 10^{-2}$ & $3.4 \times 10^{-2}$ & $7.54 \times 10^{-3}$ \\
$0.075$ &  $1.62 \times 10^{-2}$ & $1.02 \times 10^{-2}$ & $2.31 \times 10^{-3}$ \\
$0.3$ &  $4.35 \times 10^{-2}$ & $2.42 \times 10^{-2}$ & $5.29 \times 10^{-3}$
\end{tabular}
\label{tab:surrogate_errors_3D}
\end{ruledtabular}
\end{table*}

\newpage
\section{Conclusion}
\label{Sec:Conclusion}

In this study, we use the PINN method to solve a dynamic linear elasticity problem. We use state-of-the-art PINN methods and set up a deep learning framework for the same. We employ this modeling approach in three different contexts; as a forward modeling approach to solve for displacements and stresses, as an inverse modelling approach to identify unknown material properties, and finally as a surrogate forward model to identify material properties iteratively. For forward modeling, we perform a systematic hyperparameter search and we obtain an optimal set for which the total loss is minimal. We show comparison of the PINN results with respect to the FEM solutions and find that they agree qualitatively and quantitatively. It is to be noted that we use very minimal data from FEM simulations for training and stress on the analogy that this training data could be from experiments in an industrial setting. As for the inverse modeling, we show results where we obtain accurate prediction of the parameters $\lambda$ and $\mu$. However, keeping in mind the practical reasons in the industry such as minimal sensor data or infeasible measurements at the interior of the geometry, we don't delve into a deeper discussion on inverse models. Instead, we would emphasize the use of PINN in the case of creating a surrogate model. 

Initially, we show how PINN could be used as a surrogate model for a 2D dynamic problem where we accurately predict solutions for an unseen or testing value of the parameter. Later, we create a surrogate model for a 3D dynamic problem.  We show contour and line plots of the results for an unseen parameter in comparison with the FEM solution and we observe an accurate match in the displacements. The results obtained would significantly reduce the computational cost of searching for the material parameters iteratively. This is because, once the the PINN model is trained, it takes mere seconds for it to generate solutions for an unseen parameter. The deep learning framework introduced in this study to solve such dynamic problems could also be extended to solve dynamic problems beyond the linearly elastic regimes.  

For future work, we would be applying this PINN framework on problems of industrial relevance, most of which involve complex geometries and random excitation. In addition, we will also consider material non-linearity to our proposed solution approach. In summary, in this work, we have shown the successful application of a deep learning framework based on PINN to a realistic dynamic problem in solid mechanics. In doing so, we show how PINN can circumvent high computational costs associated with numerical simulations in such problems.

\section*{Acknowledgements}
The authors thank the computing resources and support provided by the CI/PDP Deep Learning Platform Team at Robert Bosch GmbH, Renningen, Germany. The authors acknowledge the help of Ponsuganth Ilangovan P for his inputs in setting up the FEM simulations. 
\newpage
\section*{Data Availability Statement}

The data that support the findings of this study are available from the corresponding author upon reasonable request.

\bibliography{Article_file}
\bibliographystyle{abbrvnat2}

\end{document}